\documentclass[10pt,twocolumn,letterpaper]{article}

\usepackage[pagenumbers]{cvpr} 

\usepackage{cuted} 

\usepackage{currfile} 

\usepackage{caption} 

\usepackage{microtype}

\usepackage{xspace}
\usepackage{multirow}
\usepackage{tabularx}
\usepackage{colortbl}
\definecolor{bestblue}{RGB}{226,238,249}
\usepackage[breakable,skins]{tcolorbox}
\tcbset{sbprompt/.style={breakable,enhanced,colback=cvprblue!3,colframe=cvprblue!55,boxrule=0.5pt,arc=1.5pt,left=4pt,right=4pt,top=3pt,bottom=3pt,fonttitle=\footnotesize\bfseries,coltitle=white,colbacktitle=cvprblue!75,fontupper=\footnotesize\raggedright,before upper={\setlength{\parskip}{0.35em}}}}

\usepackage{tikz}
\usetikzlibrary{positioning,arrows.meta,fit,calc,backgrounds,decorations.pathreplacing,shapes.geometric}

\definecolor{bpOurs}{RGB}{0,114,178}      
\definecolor{bpPrior}{RGB}{213,94,0}      
\definecolor{bpData}{RGB}{0,158,115}      
\definecolor{bpAccent}{RGB}{230,159,0}    
\definecolor{bpGray}{RGB}{110,110,110}    
\definecolor{bpLight}{RGB}{238,238,238}   
\colorlet{bpOursFill}{bpOurs!18}
\colorlet{bpPriorFill}{bpPrior!15}
\colorlet{bpDataFill}{bpData!15}

\tikzset{
  bp/base/.style   ={draw=black!70, line width=0.5pt, rounded corners=2pt, align=center,
                     font=\sffamily\scriptsize, minimum height=0.6cm, inner sep=3pt},
  bp/module/.style ={bp/base, fill=white},
  bp/frozen/.style ={bp/base, fill=bpLight, draw=bpGray, text=black!80},
  bp/ours/.style   ={bp/base, fill=bpOursFill, draw=bpOurs, line width=0.7pt},
  bp/prior/.style  ={bp/base, fill=bpPriorFill, draw=bpPrior},
  bp/data/.style   ={bp/base, fill=bpDataFill, draw=bpData},
  bp/todo/.style   ={bp/base, dashed, draw=bpGray, text=bpGray},
  bp/flow/.style   ={-{Latex[length=1.6mm]}, line width=0.6pt, black!75},
  bp/note/.style   ={font=\sffamily\tiny, text=bpGray, align=center},
  bp/panel/.style  ={font=\sffamily\footnotesize\bfseries, anchor=west},
  bp/frame/.style  ={draw=black!70, fill=white, line width=0.4pt},
}

\AtEndPreamble{%
  \crefname{figure}{Figure}{Figures}%
  \Crefname{figure}{Figure}{Figures}%
  \crefname{table}{Table}{Tables}%
  \Crefname{table}{Table}{Tables}%
}

\definecolor{cinavy}{HTML}{156082}\definecolor{cired}{HTML}{B14E3B}
\newcommand{\cibaraxis}{\begin{tikzpicture}[x=0.22cm,baseline=-0.6ex]
  \draw[gray!70,line width=0.3pt] (-4,0) -- (5,0); \draw[gray!60,dashed,line width=0.3pt] (0,-0.13) -- (0,0.13);
  \node[font=\tiny,inner sep=0,anchor=south] at (-4,0.05) {$-4$}; \node[font=\tiny,inner sep=0,anchor=south] at (0,0.1) {$0$};
  \node[font=\tiny,inner sep=0,anchor=south] at (5,0.05) {$+5$};\end{tikzpicture}}
\newcommand{\cibarbase}[4]{\begin{tikzpicture}[x=0.22cm,baseline=-0.6ex]
  \draw[gray!45,line width=0.25pt] (-4,0) -- (5,0); \draw[gray!60,dashed,line width=0.3pt] (0,-0.13) -- (0,0.13);
  \draw[line width=0.9pt,#4] (#2,0) -- (#3,0); #1\end{tikzpicture}}
\newcommand{\cibar}[4]{%
  \ifstrequal{#4}{axis}{\cibaraxis}{%
  \ifstrequal{#4}{up}{\cibarbase{\fill[cinavy] (#1,0) circle (0.08);}{#2}{#3}{cinavy}}{%
  \ifstrequal{#4}{down}{\cibarbase{\fill[cired] (#1,0) circle (0.08);}{#2}{#3}{cired}}{%
  \cibarbase{\fill[white] (#1,0) circle (0.07); \draw[gray,line width=0.5pt] (#1,0) circle (0.07);}{#2}{#3}{gray}}}}}
\definecolor{cvprblue}{rgb}{0.21,0.49,0.74}
\usepackage[breaklinks,colorlinks,allcolors=cvprblue]{hyperref}

\newcommand{\github}{\raisebox{-1.5pt}{\includegraphics[height=1.0em]{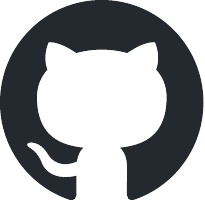}}\xspace}
\newcommand{\web}{\raisebox{-1.5pt}{\includegraphics[height=1.0em]{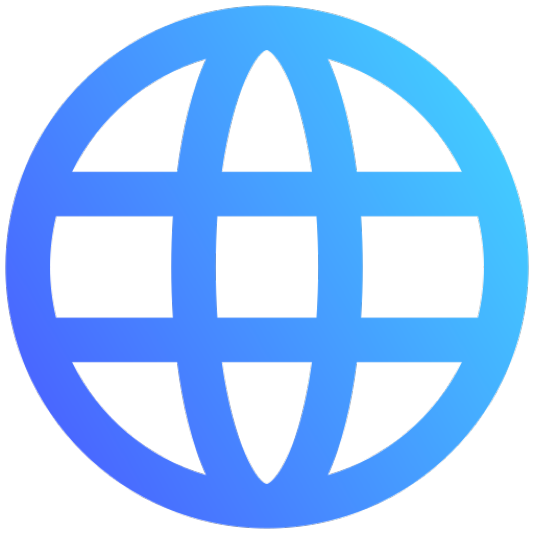}}\xspace}

\def\confName{CVPR}
\def\confYear{2027}

\title{EduGraph: A Data Infrastructure Connecting Curriculum Knowledge and Learner Evidence for Teaching Agents}

\author{Hexuan Jin$^{1}$\thanks{Equal contribution}, Xinglong Xu$^{1*}$, Yuxia Zhu$^{2*}$, Caijun Jia$^1$, Bihui Yu$^1$, Asher Qin$^{3\dagger}$, Cheng Tan$^3$\thanks{Corresponding author} \\
$^1$University of Chinese Academy of Sciences $^2$Peking University $^3$Tencent 
}

\begin{document}
\maketitle

\begin{abstract}
Teaching agents need to connect curriculum expectations with evidence of individual learning. However, curriculum standards, textbook catalogs, and learner observations are organized around different knowledge units, making their relationships difficult to establish and maintain. We present EduGraph, a data infrastructure that uses curriculum standards as the central anchor to construct a shared and versioned concept space, onto which 527 textbook volumes from 18 publishers and editions are mapped as publisher-specific views. The knowledge space spans the K–12 education system, covering 5,269 teachable concepts from primary, junior, and senior education across ten subjects. It further organizes 4,882 prerequisite relations among concepts to represent structured learning paths across educational stages. Building on this unified foundation, EduGraph grounds each learning observation in two distinct roles: the concept targeted by the question and, when the evidence supports one, the learner issue it exposes. These groundings are recorded as append-only evidence and projected into a replayable learner state. An agent-facing interface exposes curriculum knowledge and learner state to teaching agents. Through resource audits, grounding evaluations, and teaching-decision assessments, we demonstrate how EduGraph provides a reusable data foundation that agents can directly consume or integrate with existing memory.
\par\vspace{0.5em}
\noindent\makebox[\linewidth][c]{%
\github~\href{https://github.com/openraiser/EduGraph}{\textbf{Code}}\quad 
\web~\href{https://openraiser.github.io/EduGraph/}{\textbf{Website}}\quad}
\par
\end{abstract}
\vspace{-0.8em}

\section{Introduction}
\label{sec:intro}

Teaching agents must interpret learner interactions and determine how instruction should proceed, whether to continue an explanation, revisit a prerequisite, or probe understanding. Such decisions require more than the immediate response: agents need to locate the question within the broader curriculum and relate the learner's response to accumulated evidence. For example, an incorrect answer to a fraction addition problem does not necessarily reveal the underlying difficulty. The issue may concern the target concept, a prerequisite, or remain indeterminable from the available evidence. Thus, effective teaching requires connecting curricular expectations with evidence of individual learning. Recent tutoring systems have explored pedagogical planning based on demonstrations, feedback, and dialogue~\cite{liu2024socraticlm,kwon2024biped,li2026planning}, while long-term tutoring benchmarks distinguish evidence acquisition, learner-state diagnosis, and action selection~\cite{li2026longtutor}. Despite these advances, the educational data needed to support such capabilities remains fragmented and lacks a unified representation of curriculum knowledge and learner evidence.

\begin{figure*}[ht]
  \centering
  \includegraphics[width=\textwidth]{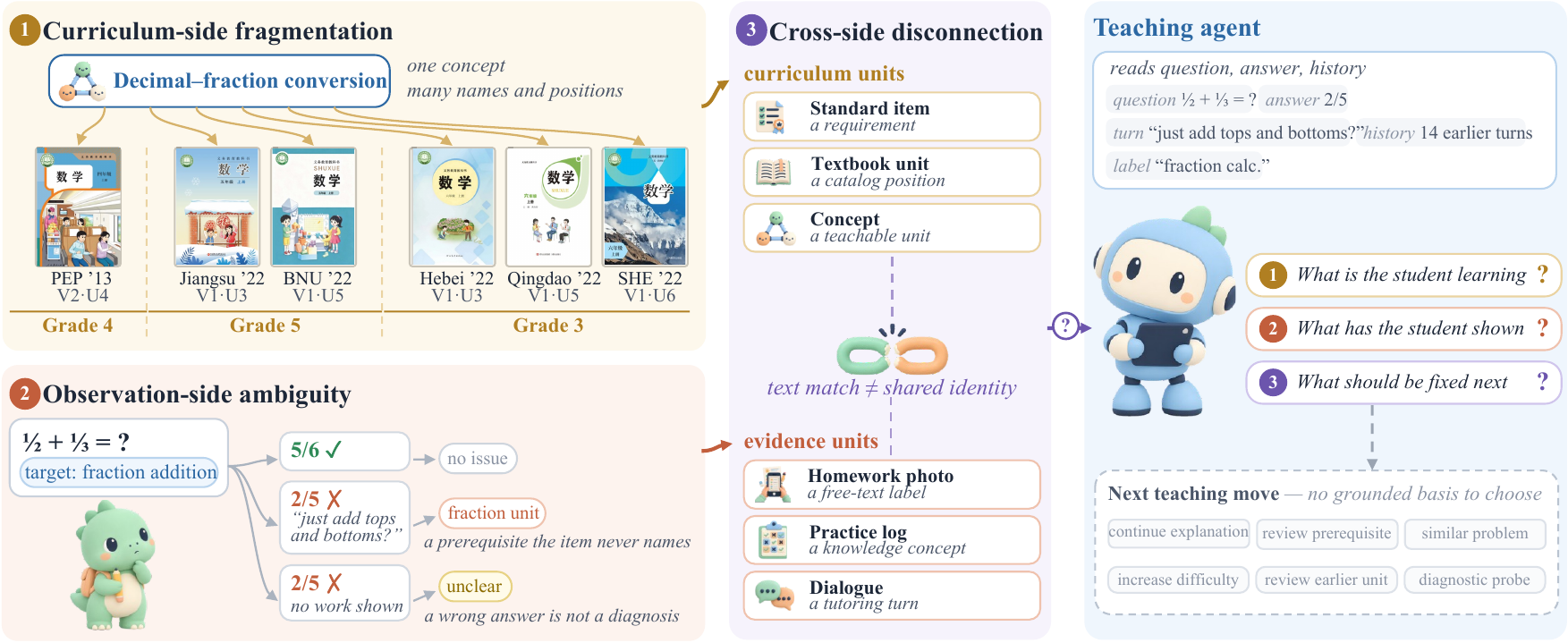}
  \caption{\textbf{Three challenges in supporting teaching agents with educational data.} (1) Curriculum-side fragmentation: Textbook editions differ in naming, organization, and grade placement. (2) Observation-side ambiguity: Learner difficulties may involve prerequisites or remain unclear. (3) Cross-side disconnection: Curriculum knowledge and learner evidence lack shared identities.}
  \label{fig:fig1_overview}
\end{figure*}

As illustrated in Figure~\ref{fig:fig1_overview}, existing educational data faces three fundamental challenges: \textbf{(1) Curriculum knowledge is fragmented across educational resources.} Although curriculum standards define what students should learn, textbooks from different publishers and editions vary in concept naming, organization, and grade placement, making it difficult to establish shared knowledge identities and learning paths. \textbf{(2) Learner observations are inherently ambiguous.} A question identifies its target concept, but the learner's actual difficulty may lie in a prerequisite or remain indeterminable from the available evidence. \textbf{(3) Curriculum knowledge and learner evidence remain disconnected.} Without a shared representation linking curricular concepts to questions and learner observations, evidence cannot be associated with the same knowledge concepts. Consequently, teaching agents lack a unified understanding of both what should be learned and what a learner has demonstrated.

We present EduGraph, a data infrastructure that connects curriculum knowledge with learner evidence through a shared, versioned concept space. EduGraph uses curriculum standards as the central anchor to organize K12 knowledge across subjects and educational stages, mapping textbook catalogs from different publishers and editions onto shared concept identities as publisher-specific views. Prerequisite relations further organize concepts into structured learning paths. Building on this foundation, EduGraph introduces dual-role grounding to distinguish the question target, which identifies what an item assesses, from the learner issue, which identifies the concept or prerequisite supported by the learner's evidence. The two roles are grounded separately, with abstention when evidence is insufficient. The resulting groundings are stored as append-only evidence and projected into a deterministic, replayable learner state. An agent-facing interface exposes curriculum position, prerequisite relations, and supporting evidence to teaching agents.

Our work bridges educational knowledge graphs and agent-based tutoring. Existing graphs organize curriculum concepts and relations~\cite{niu2025tree,liang2026k12,lu2026scimkg}, while tutoring systems use demonstrations, feedback, and dialogue for pedagogical planning~\cite{liu2024socraticlm,kwon2024biped,li2026planning}. Long-term tutoring benchmarks further emphasize the connection between evidence acquisition, learner-state diagnosis, and teaching action selection~\cite{li2026longtutor}. However, a unified data foundation connecting standards-aligned curriculum knowledge with individual learning evidence remains underexplored. EduGraph addresses this gap through a shared concept space and persistent learner evidence, providing structured context for teaching agents.

We systematically evaluate EduGraph through studies of resource quality, knowledge grounding, and agent integration. The results demonstrate its ability to organize educational resources, connect curriculum concepts with learner evidence, and provide structured context for teaching agents with or without existing memory. Together, these findings establish EduGraph as a reusable data infrastructure that teaching agents can directly consume.

Our contributions are summarized as follows:
\begin{itemize}
\item We present EduGraph, a standards-anchored K12 data infrastructure that unifies curriculum knowledge across textbook publishers and editions.
\item We introduce dual-role grounding to distinguish question targets from learner issues, with append-only evidence and replayable states connecting individual learning observations to curriculum knowledge.
\item We evaluate EduGraph across resource quality, knowledge grounding, and teaching-agent integration with different memory settings, with or without raw history.
\end{itemize}

\section{Related Work}
\label{sec:related_work}

\subsection{Knowledge Graphs and Data Infrastructure}

Knowledge graph construction organizes heterogeneous text into consistent entities and relations through information extraction, schema induction, and entity consolidation~\cite{zhang2024extract,mo2026kggen,bai2026autoschemakg}. Tree-KG uses textbook hierarchies to construct an initial graph and iteratively expands its relations~\cite{niu2025tree}.For integration across sources,ChatEA combines graph structure with language model reasoning to align equivalent entities across graphs~\cite{jiang2024unlocking}. To support graph expansion and reuse, construction and enrichment methods incorporate generation verification, expansion control, and conflict resolution to maintain the quality and consistency of new knowledge~\cite{chen2024sac,lu2026karma}. Domain knowledge graphs connect textbooks, exercises, and multimodal materials through shared concepts while encoding curricular relations and learning objectives~\cite{lu2026scimkg}. These graphs are also used to generate curriculum knowledge benchmarks, fine-tuning examples, and adaptive tutoring dialogues for evaluation and training~\cite{liang2026k12,yu2026cognet}.

For knowledge access, RAPTOR enables hierarchical retrieval~\cite{sarthi2024raptor}, while graph-based retrieval methods combine textual and relational information for question answering and external memory~\cite{he2024g,luo2026g,gutierrez2025rag}. Benchmarks evaluate joint retrieval over text and relations, as well as graph construction, retrieval, and generation pipelines, distinguishing retrieval quality from downstream performance~\cite{wu2024stark,xiang2026use}. This work focuses on mappings across resources, versioned knowledge access, and evidence traceability to provide a shared knowledge basis for maintaining learner states.

\subsection{Learner Modeling and Knowledge Tracing}

Learner modeling characterizes individual knowledge mastery and difficulties from learning records, while knowledge tracing models their evolution through interactions. Existing methods incorporate cognitive traits, prerequisite relations, and multiple levels of concept granularity~\cite{zhou2024predictive,zhang2026mgcd}. Semantic approaches combine language, behavioral representations, and cross-course concept graphs for diagnosis and knowledge transfer~\cite{dong2025knowledge,liu2025language,han2025contrastive}. Interpreting learner evidence is essential for maintaining these representations. Student questions and erroneous reasoning provide diagnostic cues, while benchmarks assess misconception understanding~\cite{kim2025knowledge,chen2026malrulelib,zhang2026misconceptionbench}. CIKT constructs learner profiles from historical performance and prediction feedback~\cite{li2025cikt}, while heterogeneous networks and learning communities support trajectory retrieval and comparison~\cite{duan2026hise,duan2026microc}. ParLD continuously updates cognitive states through dialogue feedback~\cite{yao2026conversational}.

Learner models also inform personalized teaching decisions, including review scheduling, adaptive exercise selection, and learning path planning through reinforcement learning~\cite{shu2024karl,ozyurt2026personalized}. LongTutor separately evaluates historical evidence acquisition, knowledge state diagnosis, and teaching action selection to assess how models leverage long-term learning records for personalized tutoring~\cite{li2026longtutor}. In EduGraph, we separately ground problem targets and learner issues in a shared curriculum knowledge space, retain supporting evidence and revision links for traceability, and continuously maintain individual learner states across teaching sessions.

\subsection{LLM-based Tutoring Agents}

LLM tutors develop pedagogical strategies through demonstrations, error verification, and interaction feedback~\cite{liu2024socraticlm,daheim2024stepwise}. Online reinforcement learning and preference optimization further improve teaching strategies and interaction efficiency~\cite{dinucu2025problem,shi2026beyond}. BIPED separates pedagogical action prediction from response generation, while ScaffoldLM combines stepwise planning with assessment-driven memory~\cite{kwon2024biped,li2026planning}. LearnerCoMPASS uses cognitive diagnosis and prerequisite graphs for learning path planning, while REAT retrieves relevant teaching experience~\cite{sheng2026learnercompass,zhou2026reat}. SimClass coordinates instructional interactions across classroom roles~\cite{zhang2025simulating}. 

Evaluation must distinguish tutoring output quality from subsequent learner responses. MRBench and MathTutorBench assess tutoring quality through pedagogical criteria and task capabilities~\cite{maurya2025unifying,macina2025mathtutorbench}.
LongTutor connects evidence retrieval, state diagnosis, and teaching actions in a layered evaluation framework~\cite{li2026longtutor}. 
Student simulation studies examine behavioral fidelity, dialogue generation conditioned on historical profiles, and code revisions following feedback, respectively~\cite{scarlatos2026simulated,duan2026history,niousha2026parastudent}. This work examines the role of shared knowledge and state services in selecting teaching actions, with new interactions continually updating evidence and learner states to close the tutoring loop.

\section{EduGraph}
\label{sec:method}

EduGraph connects curriculum knowledge and learner evidence in three layers. Shared Concept identities unify textbook views and prerequisite paths (\cref{sec:infra}). Role-specific grounding separates question targets from evidence-supported learner issues and accumulates a structured learner profile (\cref{sec:state}). An agent-facing interface exposes both for teaching decisions with or without raw memory, so the agent can consult both layers together (\cref{sec:policy}).

\subsection{Curriculum Knowledge Infrastructure}
\label{sec:infra}

\begin{figure*}[t]
  \centering
  \includegraphics[width=\textwidth]{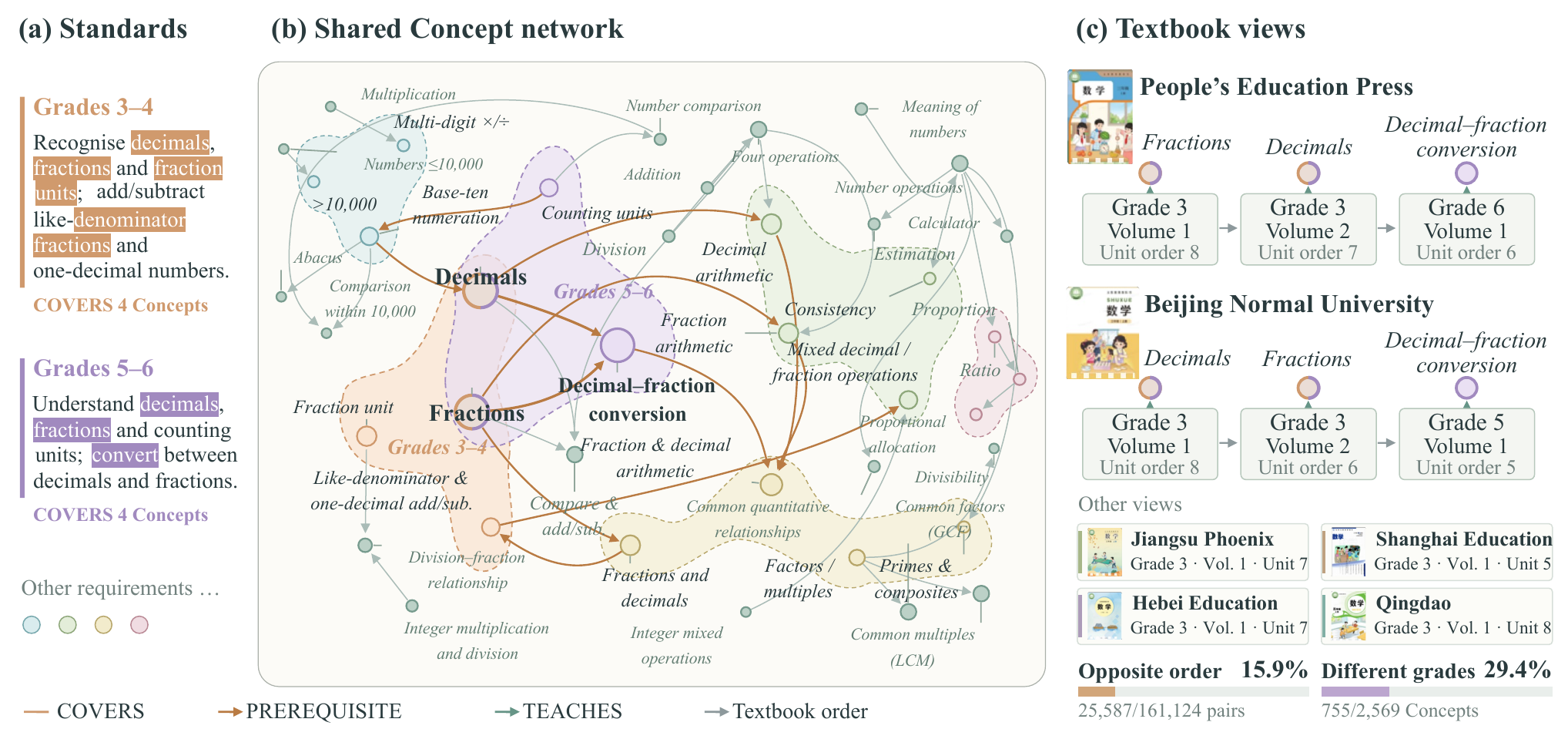}
  \caption{\textbf{Standards-anchored concept network and textbook views.}
(a) Curriculum standards define learning requirements and cover shared concepts.
(b) A unified concept network connects knowledge through prerequisite relations across grades.
(c) Publisher-specific textbook views map to shared concepts despite differences in teaching order and grade placement.}
  \label{fig:infra}
\end{figure*}

Because resources name, organize and place the same knowledge at different grades, a Concept cannot be referenced across resources by title or position alone. EduGraph anchors the knowledge space in common curriculum requirements, treats senior-school examination outlines as a supplementary alignment source, not standards, and represents each textbook as a view over shared Concepts; Figure~\ref{fig:infra} illustrates the construction. Let $S$ be standard items, $K$ Concepts, the smallest teachable, assessable and exercisable units, $T$ Topics, and $B$ textbook chapters. The conceptual core is
\begin{equation}
R=(S,\,K,\,T,\,B,\,E_{\mathrm{cov}},\,E_{\mathrm{ref}},\,E_{\mathrm{pre}},\,E_{\mathrm{tea}}),
\end{equation}
where $\mathrm{COVERS}\subseteq S\times K$ records requirements that Concepts represent, $\mathrm{REFINES}\subseteq(K\cup T)\times(K\cup T)$, $\mathrm{PREREQUISITE}\subseteq K\times K$, and $\mathrm{TEACHES}\subseteq B\times K$ with a role attribute (\textit{primary}, \textit{mentioned} or \textit{review}).

A standard item is a requirement, not a Concept or teachable unit, so \texttt{COVERS} records represented requirements without asserting total coverage. \texttt{REFINES} is a single-parent containment forest, and \texttt{PREREQUISITE} is a directed acyclic graph over Concepts; containment is never read as learning order, and a prerequisite need not contain its successor. Key points, exercises and typed facets (Skill, Problem Type, Misconception proxy) attach to Concepts as auxiliary records where supported, outside the conceptual core. Official table-of-contents chapters are mapped to existing Concepts: each \texttt{TEACHES} endpoint is checked against the chapter and Concept records, and an unsupported match stays unmapped rather than creating an edition-specific key. Grade, volume and unit position are textbook-view metadata, not attributes of the Concept itself.

Construction starts from China's 2022 compulsory-education curriculum standards~\cite{moe2022compulsory}, with senior-school examination outlines as supplementary alignment. Models propose Concepts, containment edges, prerequisite links and standard-to-Concept alignments; deterministic gates admit proposals that satisfy the schema, \texttt{REFINES} forest and \texttt{PREREQUISITE} DAG. Because Concept keys are stable and renames or merges append aliases, references from questions, exercises and evidence survive catalog edits. These identities form the shared coordinate system.

\subsection{Learner Evidence Grounding and State}
\label{sec:state}

An observation answers two questions at once. The question target is the knowledge the item was designed to assess, a property of the problem. The learner issue is the difficulty the attempt evidences, a property of the learner's evidence; it may lie in an unnamed prerequisite, or the evidence may locate nothing. A ranked list conflates them, so a salient problem word can stand in for a diagnosis.

EduGraph grounds both roles of an observation $x$---a question, a response, or dialogue---in a five-type interface vocabulary: Concept, Skill, Problem Type, Prerequisite and Misconception proxy. The core provides shared Concept identities; Skill, Problem Type and Misconception proxy are auxiliary facets with Concept associations where supported; Prerequisite denotes a Concept reached through a dependency edge. For role $r\in\{\mathrm{qt},\mathrm{li}\}$, retrieval is scoped by subject, stage and topic and combines lexical, dense, alias, key-point and exercise channels over core and auxiliary records. The target query uses the problem statement; the issue query uses the available response and dialogue, expands candidates to prerequisite ancestors and matches pitfall entries, producing a role-specific candidate set $C_r(x)$. A candidate-constrained binder $b_r(x,C_r(x))=(\sigma_r,\Sigma_r,c_r)$ returns a status $\sigma_r\in\{\textit{present},\textit{none},\textit{unclear}\}$, an entity set $\Sigma_r\subseteq C_r(x)$ of at most three, and a confidence $c_r$. Targets use Concepts, Skills or Problem Types; issues can use all five types. The binder may abstain: \textit{none} means no candidate is supported, while \textit{unclear} means the evidence cannot decide; a wrong answer alone establishes neither a misconception nor a specific prerequisite diagnosis.

Each attempt is one learner's evidence record $e=(i,x,g_{\mathrm{qt}},g_{\mathrm{li}},y,n,e_{\mathrm{sup}})$: $i$ is a unique identifier, $x$ the observation, $g_{\mathrm{qt}}$ and $g_{\mathrm{li}}$ the two role groundings, $y$ an optional graded outcome, $n$ the append position, and $e_{\mathrm{sup}}$ optionally points to an earlier record of the same learner being corrected. Records are immutable: corrections append revisions, and a revision chain that crosses learners or forms a cycle is rejected. The learner state after $n$ records is the projection
\begin{equation}
L_n=\mathrm{Project}(E_{\le n},R),\qquad E_{\le n}=\{e_1,\dots,e_n\},
\end{equation}
of one learner's evidence prefix through the graph. Project uses the effective corrected records of the prefix: a superseded value remains for audit but is not added again to current summaries. It groups records by supported Concept identities and associations into Concept profiles: observations and role groundings, target evidence, issue evidence indexed under the issue entity's own associated Concept rather than automatically under the question target, graded outcomes where available, unresolved unknowns, and supporting evidence identifiers; typed records without a supported association stay in evidence, not forced into Concept profiles. Because the projection reads only $E_{\le n}$ and $R$ under fixed settings, the same prefix and graph reproduce the same state, and rebuilding an earlier prefix never consults observations from that learner's later recorded history.

An outcome contributes to a Concept's graded-outcome summary only when the grade and that Concept are explicitly available; a common response grade is not copied to every target or issue entity. Ungraded observations add grounding and evidence, not graded-outcome estimates; corrected profiles are recomputed from valid graded evidence. A missing grade is not a failure, and performance is not inferred through prerequisite or containment neighbors; mastery is claimed only where graded evidence narrowly supports it. Concepts without graded evidence remain unknown or insufficient rather than zero mastery.

\subsection{Agent-facing Interface}
\label{sec:policy}

A shared graph and projected state still leave the agent to join them, and a teaching agent must write its next observation back into the same evidence base. EduGraph exposes the join as four calls, \texttt{search}, \texttt{ingest}, \texttt{state} and \texttt{context}, plus an optional \texttt{decide} call recording a policy's action and the evidence identifiers that it cites.

\texttt{search} returns ranked typed entities within a subject, stage or topic scope, from the conceptual core and auxiliary records. \texttt{ingest} grounds both roles of an observation, appends the resulting evidence record, and returns its identifier and groundings; appending is the only evidence write path, so history cannot be rewritten. \texttt{state} returns $L_n$ for the current evidence prefix of that learner.

\texttt{context} starts from a target $\tau$, which may be an external curriculum label, aligns it to a shared Concept with a relation $\rho\in\{\textit{exact},\textit{broader},\textit{narrower},\textit{related},\textit{unmapped}\}$, and returns a target-centered card: the aligned Concept, its prerequisite ancestors and successors, same-topic Concepts with evidence, and the learner profile restricted to those Concepts. An unmapped target returns the alignment status and no invented Concept. The card joins curriculum knowledge and learner evidence around the target while keeping them separate in the returned context.

\texttt{decide} is optional and records an action $a$ chosen by a policy, a target Concept $k$, reasons $R_d$, cited evidence identifiers $E_{\mathrm{ids}}$ and a confidence $c_d$; it does not choose the move. Recorded decisions cite evidence available to that learner at decision time, so they are traceable. Possible actions are explaining further, reviewing a prerequisite, giving a similar problem, increasing difficulty, revisiting earlier material, or posing a diagnostic probe. A host may execute the move and pass the next observation back, closing the loop Evidence $\rightarrow$ Grounding $\rightarrow$ State $\rightarrow$ Decision $\rightarrow$ New Evidence.

Two integration settings share the contract. In the direct setting, the agent answers from the target-centered card without a raw transcript. In the augmenting setting, the graph and state accompany existing raw memory, adding stable curriculum identities and an evidence-linked profile without replacing it. In both, the agent sees the same typed entities, state projection and append semantics, so a recorded decision is traceable and the next observation extends the same state through the same append path.

\section{Experiments}
\label{sec:exp}

Three questions organize the evaluation. \textbf{RQ1} asks whether the curriculum infrastructure is usable: broad, structurally sound, with preserved textbook views, coverage, and quality. \textbf{RQ2} asks whether it retrieves and binds learner evidence correctly. \textbf{RQ3} asks whether it supports teaching-agent decisions with and without raw history. RQ1 counts use release v0.21.0; RQ2 and RQ3 use the release-plus-overlay snapshot with 4,882 prerequisite edges, and RQ2's stored baselines and ablation panel use the initial 3,238-edge annotation.

\paragraph{Protocol.} Final reference labels were model-prefilled and confirmed by two teachers. RQ1 used a three-model prefill panel; RQ2 and RQ3 used a five-model panel (GPT-6.1-Sol~\cite{openai2026gpt61sol}, Claude-Opus-5.5~\cite{anthropic2026claudeopus55}, Gemini-4-Argon~\cite{deepmind2026gemini4argon}, Kimi-K3~\cite{team2026kimi}, and GLM-5.3~\cite{zeng2026glm}). Prefill agreement is $\kappa=0.479$ to $0.600$ among prefilling models, not teachers; all reported values use stored references and splits.

\subsection{RQ1: Resource Usability}
\label{sec:rq1}
\label{sec:e1}

\begin{figure*}[t]
  \centering
  \includegraphics[width=\textwidth]{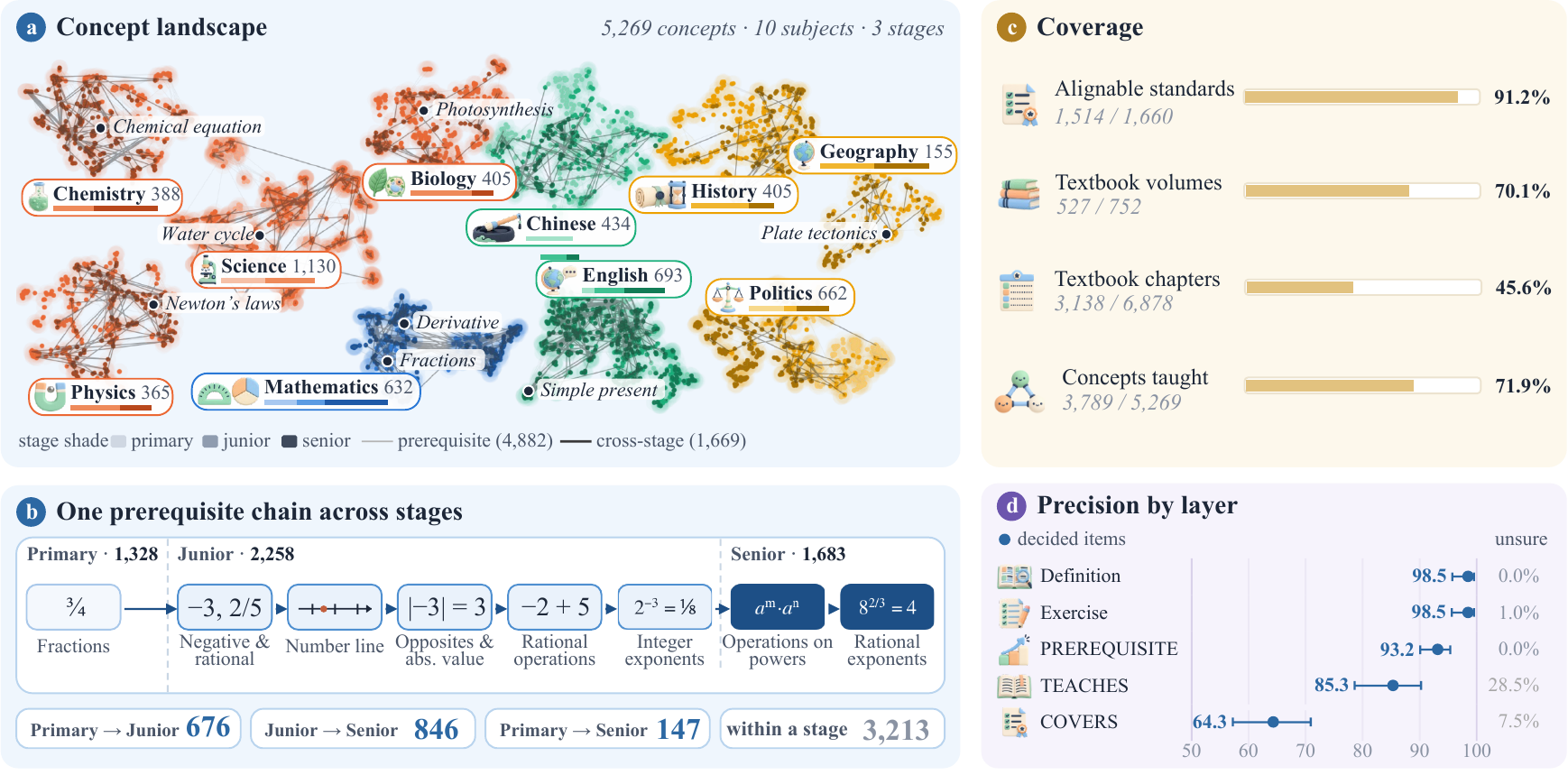}
  \caption{\textbf{EduGraph knowledge structure, coverage, and quality.} (a) Concept landscape showing 5,269 concepts across ten subjects and three educational stages, connected by prerequisite relations. (b) An illustrative prerequisite chain spanning primary, junior, and senior education, with cross-stage relation counts. (c) Coverage of curriculum standards, textbook volumes, chapters, and concepts. (d) Teacher-confirmed precision by data layer with 95\% confidence intervals; unsure cases are excluded from precision and reported separately.}
  \label{fig:resource}
\end{figure*}

\textbf{Breadth.} EduGraph spans ten subjects and the primary, junior-secondary, and senior-secondary stages (\cref{fig:resource}a), with 35,937 major resource records. Its 5,269 Concepts and 1,322 Topics form the knowledge core, and KeyPoints and Exercises extend the index to explanations, methods, and practice, with Concept associations where supported.

\noindent\textbf{Structure.} \texttt{REFINES} organizes content containment and \texttt{PREREQUISITE} learning dependencies; the 4,882 prerequisite edges form a directed acyclic graph whose longest path contains 12 Concepts across all three stages (\cref{fig:resource}b).

\noindent\textbf{Textbook views.} Among eligible Concept pairs co-taught in a grade, 15.9\% appear in opposite orders across publishers, and 29.4\% of multi-publisher Concepts have different first teaching grades. \texttt{COVERS} links standard items to multiple Concepts, while \texttt{TEACHES} maps chapters onto the same nodes retaining grades and order (\cref{fig:infra}c); identifiers stay stable across editions. Prerequisite direction matches textbook order in 78.3\% of comparable same-stage observations, 28 points above random pairs and \texttt{REFINES} siblings.

\noindent\textbf{Coverage.} Internally, 91.2\% of alignable standard items have a \texttt{COVERS} link, 71.9\% of Concepts link to textbook chapters, and 70.1\% of volumes and 45.6\% of chapters carry a \texttt{TEACHES} edge (\cref{fig:resource}c).

\noindent\textbf{Quality.} Teacher-confirmed audits give 98.5\% precision for definitions; 98.5\% decided and 97.5\% conservative for exercises; 93.2\% for prerequisites (95\% CI 90.1 to 95.3); 85.3\% decided and 61.0\% conservative for \texttt{TEACHES} with 28.5\% abstention; and 64.3\% decided and 59.5\% conservative for \texttt{COVERS}, including 19.4\% for English and 16.7\% for geography (\cref{fig:resource}d). These are per-layer estimates on different populations, not one overall score; the supplement reports the prerequisite sensitivity as 114/122 $=$ 93.4\%.

\subsection{RQ2: Evidence Grounding}
\label{sec:rq2}
\label{sec:e2}

RQ2 contains 1,000 real observations (200 development, 800 test, 956 groups) with shuffled pools of 36 to 57 (mean 48.6). Task A retrieves over the full 5,269-Concept catalog: Skill and Misconception-proxy key points project to their parent Concept, prerequisite proxies keep Concept IDs, and Problem Type labels are excluded because they lack a Concept target. Against the leave-one-out GPT-6.1-Sol reference (635 question targets, 117 learner issues), hierarchical retrieval reaches .437 and .462, above the strongest routed baseline. Tuned fusion reaches .809 and .756, and equal-weight fusion .793 and .782, well above the LLM ranker on learner issues. The graph alone trails LLM-direct on question targets (.437 versus .608) and is comparable on learner issues (.462 versus .436, a paired bootstrap interval crossing zero).

\begin{table}[t]
\centering
\footnotesize
\caption{\textbf{Retrieval and role binding fail differently, and EduGraph addresses both.} (a) Recall@10 on the RQ2 test cases with a question target ($n=635$) or learner issue ($n=117$). (b) Binding on the fixed candidate pool, mean over five backbones. Best in bold; metric definitions are in the text.}
\label{tab:e2}
{\setlength{\tabcolsep}{4pt}%
\begin{tabular*}{\linewidth}{>{\raggedright\arraybackslash}p{\dimexpr0.58\linewidth-2\tabcolsep\relax}>{\centering\arraybackslash}p{\dimexpr0.21\linewidth-2\tabcolsep\relax}>{\centering\arraybackslash}p{\dimexpr0.21\linewidth-2\tabcolsep\relax}}
\toprule
\multicolumn{3}{@{}l}{\textit{(a) Retrieval, Recall@10}}\\
Method & QT & LI \\
\midrule
Routed hybrid + reranker & .404 & .373 \\
\rowcolor{bestblue} EduGraph hierarchical & .437 & .462 \\
LLM-direct, 3 labels & .608 & .436 \\
LLM-ranked top-10 & .771 & .564 \\
\rowcolor{bestblue} \quad + LLM fusion, tuned & \textbf{.809} & .756 \\
\rowcolor{bestblue} \quad + LLM fusion, equal RRF & .793 & \textbf{.782} \\
\bottomrule
\end{tabular*}}
\vspace{0.3em}

{\setlength{\tabcolsep}{4pt}%
\begin{tabular*}{\linewidth}{>{\raggedright\arraybackslash}p{\dimexpr0.44\linewidth-2\tabcolsep\relax}>{\centering\arraybackslash}p{\dimexpr0.19\linewidth-2\tabcolsep\relax}>{\centering\arraybackslash}p{\dimexpr0.19\linewidth-2\tabcolsep\relax}>{\centering\arraybackslash}p{\dimexpr0.18\linewidth-2\tabcolsep\relax}}
\toprule
\multicolumn{4}{@{}l}{\textit{(b) Binding, fixed candidate pool}}\\
Binder & QT F1 & LI F1 & FP \\
\midrule
Single label, no roles & .515 & .231 & .957 \\
Two roles, names only & .593 & .490 & .031 \\
\rowcolor{bestblue} EduGraph binder & \textbf{.744} & \textbf{.563} & \textbf{.029} \\
\bottomrule
\end{tabular*}}

\end{table}

Task B is entity-ID micro-F1 over the fixed pool. Across five backbones, the EduGraph binder reaches .744 and .563, above a single-label model and a names-only role-separated model (\cref{tab:e2}b). On reference-none learner-issue cases the single-label model emits an issue in 95.7\%, the binder in 2.9\%, and names-only in 3.1\%. The typed view adds .151 and .073 over names-only in an observed comparison, not a controlled causal ablation: the binder uses 4,882 edges while stored baselines use 3,238. Removing abstention drops learner-issue F1 from .475 to .238 on GPT-6.1-Sol and .568 to .311 on DeepSeek-V4.1-Flash; removing prerequisite or misconception annotations changes F1 by less than .04. On the ablation panel, densification left learner-issue R@10 at .472 and lowered MRR; the unified view is adopted for coverage, not measured retrieval gain.


\begin{figure*}[t]
  \centering
  \includegraphics[width=\textwidth]{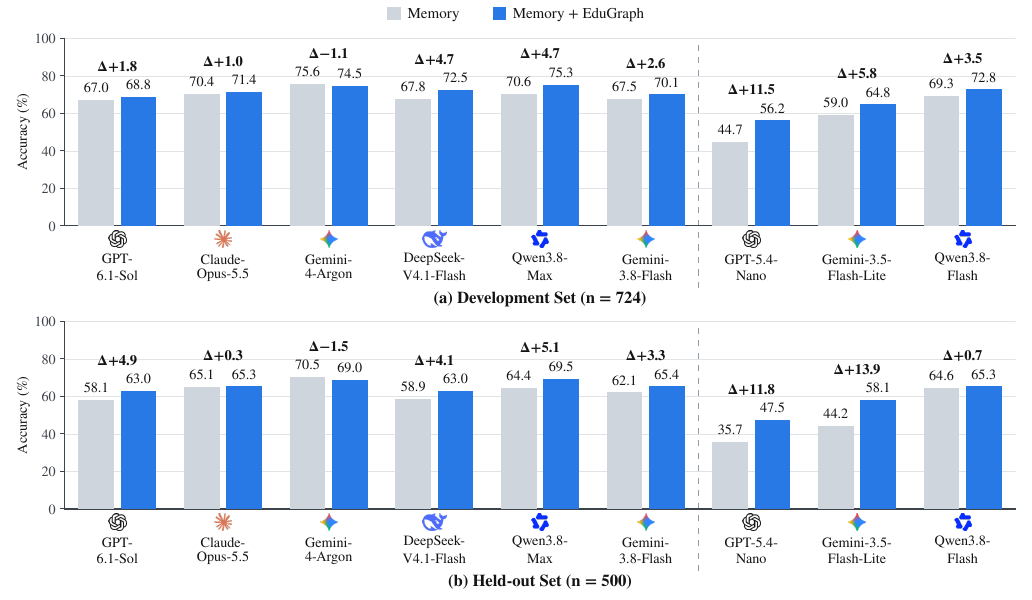}
\caption{\textbf{EduGraph improves teaching-action agreement across model backbones.}
Accuracy of Memory and Memory + EduGraph on (a) the development set (724 cases) and (b) the held-out set (500 cases), across six large and three compact models. $\Delta$ denotes the accuracy difference in percentage points; higher values indicate closer agreement with the stored evaluation references.}
  \label{fig:e3-backbones}
\end{figure*}

\subsection{RQ3: Teaching Decisions}
\label{sec:rq3}
\label{sec:e3}
Raw history is the strongest core context; adding EduGraph raises average teacher-action agreement, but unevenly across backbones. RQ3 evaluates six large backbones (GPT-6.1-Sol, Claude-Opus-5.5, Gemini-4-Argon, DeepSeek-V4.1-Flash, Qwen3.8-Max, and Gemini-3.8-Flash) and three compact ones (GPT-5.4-Nano, Gemini-3.5-Flash-Lite, and Qwen3.8-Flash) with three seeds each. The \emph{Development Set} contains 724 real decision points, 500 historical decision trajectories and 224 tutoring-dialogue turns, and is descriptive because its cases had been inspected during prompt design. The \emph{Held-out Set} contains 500 decision points from 500 distinct primary-mathematics learners, opened after the call was frozen and pre-registered. Trajectory inputs contain up to 30 real attempts and a knowledge-component summary; dialogue cases end at the student turn without the assistant answer. The reference is a five-model panel prompted as senior teachers and confirmed by two teachers. Panel members evaluated as backbones are scored against the leave-one-out plurality of the other four; non-panel backbones use the full panel. An audit found that the leave-one-out references used for evaluated panel members differ from the final teacher-confirmed labels on some cases; those accuracy values therefore use a model-specific reference and were not rescored. Vanilla sees the observation and goal; RAG adds the top five EduGraph passages; Memory adds raw within-session history; EduGraph+State without raw history adds structured state and the target prerequisite subgraph; Memory + EduGraph adds the position card and state to the same raw history and goal (\cref{fig:e3-backbones}).

On the Development Set, Memory + EduGraph improves average accuracy over Memory by 2.3 points for large models and 6.9 for compact models. On the Held-out Set, the corresponding gains are 2.7 points (95\% CI 0.6 to 5.0) and 8.8 points (6.3 to 11.2; $p<.001$). The large-model held-out gain is positive by bootstrap interval but not significant under the exact McNemar test at $p=.076$.

The averages hide real heterogeneity: eight of nine backbones improve accuracy on each set, while Gemini-4-Argon declines. Without raw history, EduGraph+State raises compact-model accuracy by 3.7 points on the Development Set but leaves large-model accuracy 2.7 points below Memory, which uses raw history (McNemar $p=.143$).

Teacher-action agreement is not a learning outcome. Review is too rare for a stable per-action estimate (one case in each set). These results therefore describe agreement under the stored evaluation references, not a measured learning outcome for any student in either set.

\begin{figure*}[t]
  \centering
  \includegraphics[width=\textwidth]{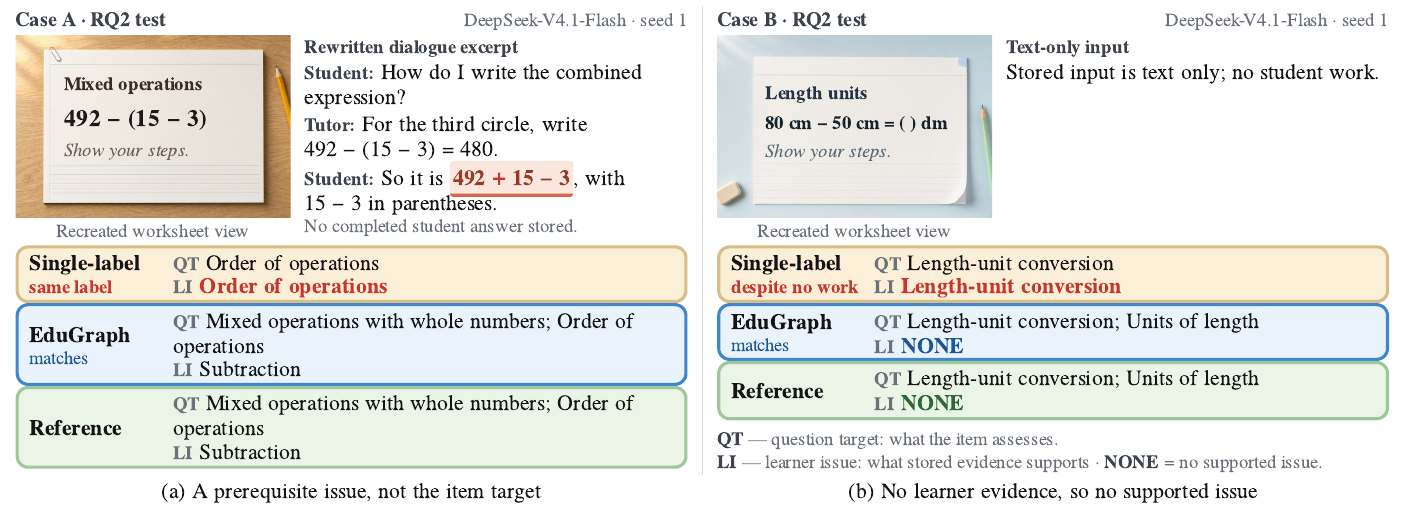}
  \caption{\textbf{Restatements can expose a hidden issue.} (a) The rewritten English dialogue exposes an outer-operator error. (b) No student work supports a unit-conversion issue. Worksheet views are generated illustrations of stored text-only inputs. Blue: EduGraph; mint: reference; coral: unsupported binding or the student's operator error shown in the rewritten dialogue.}
  \label{fig:cases-rq2}
\end{figure*}

\begin{figure*}[t]
  \centering
  \includegraphics[width=\textwidth]{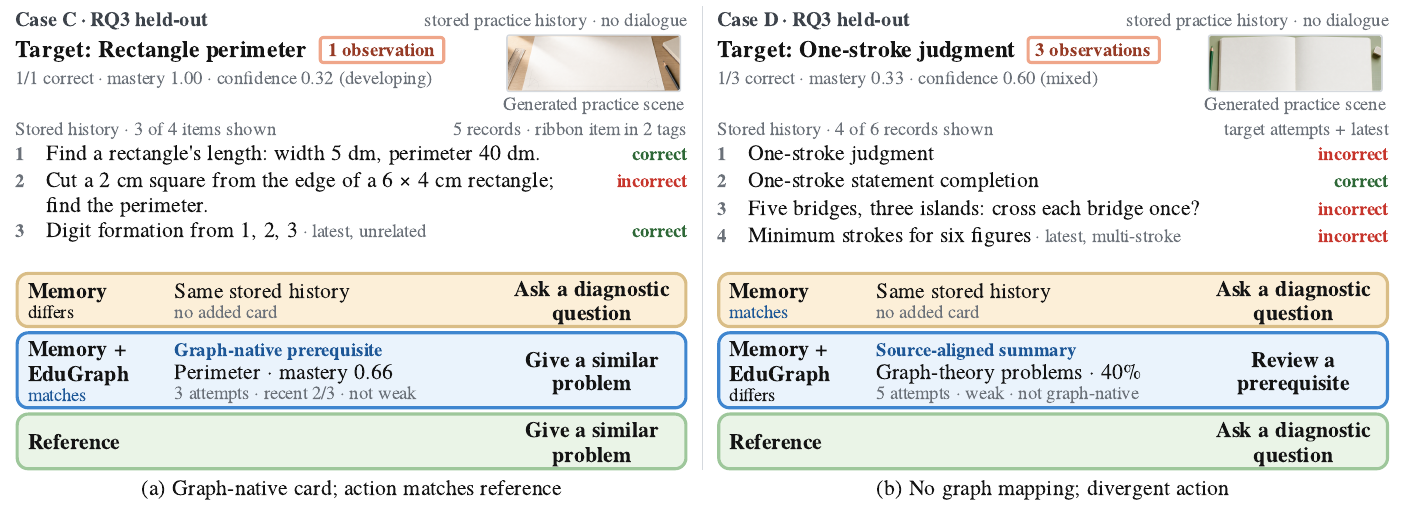}
  \caption{\textbf{Curriculum context can help or mislead.} Generated practice scenes accompany identical stored histories in both arms. (a) C maps broadly to rectangle perimeter and area formulas; its added card contains a native prerequisite. (b) D is unmapped; its source-aligned summary supplies no native edge. No learner dialogue is stored. Blue: Memory + EduGraph; butter: Memory; mint: reference.}
  \label{fig:cases-rq3}
\end{figure*}

\subsection{Case Studies}
\label{sec:cases}

\Cref{fig:cases-rq2} shows why one label cannot support instruction. The student's outer-operator error sits in a prerequisite, while the item targets mixed operations. EduGraph returns both roles separately; the baseline collapses them. When no student work exists, only abstention is evidence-safe, yet the baseline still invents an unsupported issue.

\Cref{fig:cases-rq3} shows value and failure. In C, a graph-native prerequisite lets Memory + EduGraph choose the similar-problem reference; Memory asks a diagnostic question. In D, a source-aligned summary for an unmapped target triggers prerequisite review, while Memory follows the diagnostic reference. The interface supports useful adaptation, yet can also produce unsupported remediation when mappings fail.

\section{Conclusion}
\label{sec:conclusion}

EduGraph treats curriculum structure and learner evidence as one shared teaching substrate. Standards anchor concept identity, textbook catalogs expose alternative curricula, and role-specific grounding distinguishes an item's target from the issue exposed by a learner. Append-only evidence yields a reproducible state for downstream agents. Experiments show closer agreement with teacher-confirmed decisions, especially for compact models, while the large-model result remains inconclusive. Cases also expose a boundary: source-aligned summaries can support useful actions without graph-native prerequisites, and missing mappings can cause over-remediation. Broader coverage and evaluation of student learning in interactive settings remain future work.

\clearpage
\section*{LLM Usage Statement}

OpenAI Codex was used to assist with manuscript preparation,
including drafting and revising the conclusion, adapting the
existing \LaTeX{} project to the CVPR template, formatting
tables and figures, and resolving \LaTeX{} compatibility issues.

{
    \small
    \setlength{\bibsep}{0pt}
    \bibliographystyle{ieeenat_fullname}
    \bibliography{main}

@String(AAAI = {AAAI})

@inproceedings{zhang2024extract,
  title={Extract, define, canonicalize: An llm-based framework for knowledge graph construction},
  author={Zhang, Bowen and Soh, Harold},
  booktitle={Proceedings of the 2024 conference on empirical methods in natural language processing},
  pages={9820--9836},
  year={2024}
}

@article{mo2026kggen,
  title={Kggen: Extracting knowledge graphs from plain text with language models},
  author={Mo, Belinda and Yu, Kyssen and Kazdan, Joshua and Mpala, Proud and Yu, Lisa and Kanatsoulis, Charilaos and Koyejo, Sanmi},
  journal={Advances in Neural Information Processing Systems},
  volume={38},
  pages={30092--30115},
  year={2026}
}

@inproceedings{bai2026autoschemakg,
  title={Autoschemakg: Autonomous knowledge graph construction through dynamic schema induction from web-scale corpora},
  author={Bai, Jiaxin and Fan, Wei and Hu, Qi and Zong, Qing and Li, Chunyang and Tsang, Hong Ting and Luo, Hongyu and Yim, Yauwai and Huang, Haoyu and Zhou, Xiao and others},
  booktitle={Proceedings of the 64th Annual Meeting of the Association for Computational Linguistics (Volume 1: Long Papers)},
  pages={20557--20584},
  year={2026}
}

@inproceedings{niu2025tree,
  title={Tree-KG: an expandable knowledge graph construction framework for knowledge-intensive domains},
  author={Niu, Songjie and Yang, Kaisen and Zhao, Rui and Liu, Yichao and Li, Zonglin and Wang, Hongning and Chen, Wenguang},
  booktitle={Proceedings of the 63rd Annual Meeting of the Association for Computational Linguistics (Volume 1: Long Papers)},
  pages={18516--18529},
  year={2025}
}

@inproceedings{jiang2024unlocking,
  title={Unlocking the power of large language models for entity alignment},
  author={Jiang, Xuhui and Shen, Yinghan and Shi, Zhichao and Xu, Chengjin and Li, Wei and Li, Zixuan and Guo, Jian and Shen, Huawei and Wang, Yuanzhuo},
  booktitle={Proceedings of the 62nd Annual Meeting of the Association for Computational Linguistics (Volume 1: Long Papers)},
  pages={7566--7583},
  year={2024}
}

@inproceedings{chen2024sac,
  title={SAC-KG: exploiting large language models as skilled automatic constructors for domain knowledge graph},
  author={Chen, Hanzhu and Shen, Xu and Lv, Qitan and Wang, Jie and Ni, Xiaoqi and Ye, Jieping},
  booktitle={Proceedings of the 62nd Annual Meeting of the Association for Computational Linguistics (Volume 1: Long Papers)},
  pages={4345--4360},
  year={2024}
}

@article{lu2026karma,
  title={Karma: Leveraging multi-agent llms for automated knowledge graph enrichment},
  author={Lu, Yuxing and Wu, Wei and Zhao, Xukai and Peng, Rui and Wang, Jinzhuo},
  journal={Advances in Neural Information Processing Systems},
  volume={38},
  pages={56421--56453},
  year={2026}
}

@inproceedings{lu2026scimkg,
  title={SciMKG: a multimodal knowledge graph for science education with text, image, video and audio},
  author={Lu, Tong and Wang, Zhichun and Zhou, Yaoyu and Guan, Yiming and Bai, Zhiyong and Du, Junsheng},
  booktitle={Proceedings of the AAAI Conference on Artificial Intelligence},
  volume={40},
  number={18},
  pages={15466--15474},
  year={2026}
}

@article{liang2026k12,
  title={K12-KGraph: A Curriculum-Aligned Knowledge Graph for Benchmarking and Training Educational LLMs},
  author={Liang, Hao and Lin, Qihan and Han, Zhaoyang and Ma, Xiaochen and Wong, Zhen Hao and Qiang, Meiyi and Sun, Linzhuang and Zhang, Wentao},
  journal={arXiv preprint arXiv:2605.09635},
  year={2026}
}

@inproceedings{yu2026cognet,
  title={CogNet-KG: Empowering Tutoring Dialogues with a Cognitively-Structured Knowledge Graph for STEM Learning},
  author={Yu, Ding and Lu, Yu and Li, Tengju and Xiong, Shasha and Yu, Shengquan},
  booktitle={Findings of the Association for Computational Linguistics: ACL 2026},
  pages={13101--13121},
  year={2026}
}

@inproceedings{sarthi2024raptor,
  title={Raptor: Recursive abstractive processing for tree-organized retrieval},
  author={Sarthi, Parth and Abdullah, Salman and Tuli, Aditi and Khanna, Shubh and Goldie, Anna and Manning, Christopher},
  booktitle={International Conference on Learning Representations},
  volume={2024},
  pages={32628--32649},
  year={2024}
}

@article{he2024g,
  title={G-retriever: Retrieval-augmented generation for textual graph understanding and question answering},
  author={He, Xiaoxin and Tian, Yijun and Sun, Yifei and Chawla, Nitesh V and Laurent, Thomas and LeCun, Yann and Bresson, Xavier and Hooi, Bryan},
  journal={Advances in Neural Information Processing Systems},
  volume={37},
  pages={132876--132907},
  year={2024}
}

@inproceedings{luo2026g,
  title={G-reasoner: Foundation models for unified reasoning over graph-structured knowledge},
  author={Luo, Linhao and Zhao, Zicheng and Liu, Junnan and Qiu, Zhangchi and Dong, Junnan and Panev, Serge and Gong, Chen and Vu, Thuy-Trang and Haffari, Gholamreza and Phung, Dinh and others},
  booktitle={International Conference on Learning Representations},
  volume={2026},
  pages={63769--63793},
  year={2026}
}

@article{gutierrez2025rag,
  title={From rag to memory: Non-parametric continual learning for large language models},
  author={Guti{\'e}rrez, Bernal Jim{\'e}nez and Shu, Yiheng and Qi, Weijian and Zhou, Sizhe and Su, Yu},
  journal={arXiv preprint arXiv:2502.14802},
  year={2025}
}

@article{wu2024stark,
  title={Stark: Benchmarking llm retrieval on textual and relational knowledge bases},
  author={Wu, Shirley and Zhao, Shiyu and Yasunaga, Michihiro and Huang, Kexin and Cao, Kaidi and Huang, Qian and Ioannidis, Vassilis N and Subbian, Karthik and Zou, James and Leskovec, Jure},
  journal={Advances in Neural Information Processing Systems},
  volume={37},
  pages={127129--127153},
  year={2024}
}

@inproceedings{xiang2026use,
  title={When to use graphs in rag: A comprehensive analysis for graph retrieval-augmented generation},
  author={Xiang, Zhishang and Wu, Chuanjie and Zhang, Qinggang and Chen, Shengyuan and Hong, Zijin and Huang, Xiao and Su, Jinsong},
  booktitle={International Conference on Learning Representations},
  volume={2026},
  pages={66145--66178},
  year={2026}
}

@inproceedings{zhou2024predictive,
  title={Predictive, scalable and interpretable knowledge tracing on structured domains},
  author={Zhou, Hanqi and Bamler, Robert and Wu, Charley and Tejero-Cantero, {\'A}lvaro},
  booktitle={International Conference on Learning Representations},
  volume={2024},
  pages={35609--35622},
  year={2024}
}

@article{zhang2026mgcd,
  title={MGCD: Multiple-Granularity Cognitive Diagnosis in Intelligent Education Systems},
  author={Zhang, Yuhong and He, Tiancheng and Bu, Chenyang and Yu, Kui and Hu, Xuegang and Wu, Xindong},
  journal={ACM Transactions on Information Systems},
  volume={44},
  number={5},
  pages={1--23},
  year={2026},
  publisher={ACM, New York, NY}
}

@inproceedings{dong2025knowledge,
  title={Knowledge is power: Harnessing large language models for enhanced cognitive diagnosis},
  author={Dong, Zhiang and Chen, Jingyuan and Wu, Fei},
  booktitle={Proceedings of the AAAI Conference on Artificial Intelligence},
  volume={39},
  number={1},
  pages={164--172},
  year={2025}
}

@inproceedings{liu2025language,
  title={Language representation favored zero-shot cross-domain cognitive diagnosis},
  author={Liu, Shuo and Zhou, Zihan and Liu, Yuanhao and Zhang, Jing and Qian, Hong},
  booktitle={Proceedings of the 31st ACM SIGKDD Conference on Knowledge Discovery and Data Mining V. 1},
  pages={836--847},
  year={2025}
}

@article{han2025contrastive,
  title={Contrastive cross-course knowledge tracing via concept graph guided knowledge transfer},
  author={Han, Wenkang and Lin, Wang and Hu, Liya and Dai, Zhenlong and Zhou, Yiyun and Li, Mengze and Liu, Zemin and Yao, Chang and Chen, Jingyuan},
  journal={arXiv preprint arXiv:2505.13489},
  year={2025}
}

@inproceedings{kim2025knowledge,
  title={Knowledge tracing in programming education integrating students’ questions},
  author={Kim, Doyoun and Kim, Suin and Jo, Yohan},
  booktitle={Proceedings of the 63rd Annual Meeting of the Association for Computational Linguistics (Volume 1: Long Papers)},
  pages={27703--27718},
  year={2025}
}

@inproceedings{chen2026malrulelib,
  title={Malrulelib: large-scale executable misconception reasoning with step traces for modeling student thinking in mathematics},
  author={Chen, Xinghe and Liu, Naiming and Sonkar, Shashank},
  booktitle={Proceedings of the 64th Annual Meeting of the Association for Computational Linguistics (Volume 1: Long Papers)},
  pages={15112--15138},
  year={2026}
}

@inproceedings{zhang2026misconceptionbench,
  title={MisconceptionBench: Evaluating LLMs on Science Misconceptions and Student Response Distributions},
  author={Zhang, Ying and Wang, Zhaoji and Zhang, Chenyu and Gao, Yizhu and Ma, Wenchao and Zhai, Xiaoming},
  booktitle={Proceedings of the 2026 Conference on Empirical Methods in Natural Language Processing},
  pages={16525--16539},
  year={2026}
}

@inproceedings{li2025cikt,
  title={CIKT: A collaborative and iterative knowledge tracing framework with large language models},
  author={Li, Runze and Wu, Siyu and Wang, Jun and Zhang, Wei},
  booktitle={Proceedings of the 2025 Conference on Empirical Methods in Natural Language Processing},
  pages={19332--19345},
  year={2025}
}

@inproceedings{yao2026conversational,
  title={Conversational learning diagnosis via reasoning multi-turn interactive learning},
  author={Yao, Fangzhou and Chang, Sheng and Gao, Weibo and Liu, Qi},
  booktitle={Proceedings of the AAAI Conference on Artificial Intelligence},
  volume={40},
  number={40},
  pages={34387--34395},
  year={2026}
}

@inproceedings{duan2026hise,
  title={Hise-kt: Synergizing heterogeneous information networks and llms for explainable knowledge tracing with meta-path optimization},
  author={Duan, Zhiyi and Shi, Zixing and Yuan, Hongyu and Wang, Qi},
  booktitle={Proceedings of the AAAI Conference on Artificial Intelligence},
  volume={40},
  number={17},
  pages={14693--14701},
  year={2026}
}

@inproceedings{duan2026microc,
  title={MicroC-KT: Modeling Community Effect via Learning Micro-Environment for Evidence-Grounded Explainable Knowledge Tracing},
  author={Duan, Zhiyi and Shi, Zixing and Jia, Bing and Wang, Qi},
  booktitle={Proceedings of the 64th Annual Meeting of the Association for Computational Linguistics (Volume 1: Long Papers)},
  pages={25781--25803},
  year={2026}
}

@inproceedings{shu2024karl,
  title={Karl: Knowledge-aware retrieval and representations aid retention and learning in students},
  author={Shu, Matthew and Balepur, Nishant and Feng, Shi and Boyd-Graber, Jordan Lee},
  booktitle={Proceedings of the 2024 Conference on Empirical Methods in Natural Language Processing},
  pages={14161--14178},
  year={2024}
}

@article{ozyurt2026personalized,
  title={Personalized exercise recommendation with semantically-grounded knowledge tracing},
  author={Ozyurt, Yilmazcan and Almaci, Tunaberk and Feuerriegel, Stefan and Sachan, Mrinmaya},
  journal={Advances in Neural Information Processing Systems},
  volume={38},
  pages={13019--13059},
  year={2026}
}

@inproceedings{li2026longtutor,
  title={LongTutor: Benchmarking Large Language Models for Long-term Personalized Tutoring},
  author={Li, Ning and Zhang, Zheng and Huang, Zhenya and Li, Rui and Zhan, Yi and Luo, Yinbo and Liu, Qi and Chen, Enhong},
  booktitle={Proceedings of the 64th Annual Meeting of the Association for Computational Linguistics (Volume 1: Long Papers)},
  pages={29712--29737},
  year={2026}
}

@article{liu2024socraticlm,
  title={SocraticLM: Exploring socratic personalized teaching with large language models},
  author={Liu, Jiayu and Huang, Zhenya and Xiao, Tong and Sha, Jing and Wu, Jinze and Liu, Qi and Wang, Shijin and Chen, Enhong},
  journal={Advances in Neural Information Processing Systems},
  volume={37},
  pages={85693--85721},
  year={2024}
}

@inproceedings{daheim2024stepwise,
  title={Stepwise verification and remediation of student reasoning errors with large language model tutors},
  author={Daheim, Nico and Macina, Jakub and Kapur, Manu and Gurevych, Iryna and Sachan, Mrinmaya},
  booktitle={Proceedings of the 2024 Conference on Empirical Methods in Natural Language Processing},
  pages={8386--8411},
  year={2024}
}

@inproceedings{dinucu2025problem,
  title={From problem-solving to teaching problem-solving: Aligning llms with pedagogy using reinforcement learning},
  author={Dinucu-Jianu, David and Macina, Jakub and Daheim, Nico and Hakimi, Ido and Gurevych, Iryna and Sachan, Mrinmaya},
  booktitle={Proceedings of the 2025 Conference on Empirical Methods in Natural Language Processing},
  pages={272--292},
  year={2025}
}

@inproceedings{shi2026beyond,
  title={Beyond Pedagogical Principles: Multi-Horizon Preference Optimization for Efficient Socratic Tutoring},
  author={Shi, Xin and Zhang, Chao and Zhu, Yifan and Zhang, Xueqiao and Luo, Yawei},
  booktitle={Proceedings of the 64th Annual Meeting of the Association for Computational Linguistics (Volume 1: Long Papers)},
  pages={11289--11306},
  year={2026}
}

@inproceedings{kwon2024biped,
  title={BIPED: Pedagogically informed tutoring system for ESL education},
  author={Kwon, Soonwoo and Kim, Sojung and Park, Minju and Lee, Seunghyun and Kim, Kyuseok},
  booktitle={Proceedings of the 62nd Annual Meeting of the Association for Computational Linguistics (Volume 1: Long Papers)},
  pages={3389--3414},
  year={2024}
}

@inproceedings{li2026planning,
  title={Planning-Guided Tutoring with Assessment-Driven Memory for Pedagogical LLM Tutors},
  author={Li, Zechen and Zhu, Qiannan and Wang, Mei and Li, Jia and Huang, Hua},
  booktitle={Proceedings of the 64th Annual Meeting of the Association for Computational Linguistics (Volume 1: Long Papers)},
  pages={7165--7188},
  year={2026}
}

@inproceedings{sheng2026learnercompass,
  title={LearnerCoMPASS: Intelligent Tutoring System with Dynamic Cognitive Diagnosis and Multi-Model Path Planning},
  author={Sheng, Ziji and Tie, Guiyao and Wang, Weidong and Zhou, Pan and Liu, Daizong},
  booktitle={Proceedings of the 64th Annual Meeting of the Association for Computational Linguistics (Volume 1: Long Papers)},
  pages={9016--9042},
  year={2026}
}

@inproceedings{zhou2026reat,
  title={REAT: A Reflective Experience-Augmented Tutoring Framework for Multi-turn Mathematical Instruction},
  author={Zhou, Jianheng and Zhang, Chaoli and Wei, Xingjun and Zhou, Xinliang and Fortino, Giancarlo and Fan, Xing and Wang, Yanfeng and Wen, Qingsong and Li, Haoyang},
  booktitle={Proceedings of the 2026 Conference on Empirical Methods in Natural Language Processing},
  pages={13945--13967},
  year={2026}
}

@inproceedings{zhang2025simulating,
  title={Simulating classroom education with llm-empowered agents},
  author={Zhang, Zheyuan and Zhang-Li, Daniel and Yu, Jifan and Gong, Linlu and Zhou, Jinchang and Hao, Zhanxin and Jiang, Jianxiao and Cao, Jie and Liu, Huiqin and Liu, Zhiyuan and others},
  booktitle={Proceedings of the 2025 conference of the nations of the americas chapter of the association for computational linguistics: Human language technologies (volume 1: Long papers)},
  pages={10364--10379},
  year={2025}
}

@inproceedings{maurya2025unifying,
  title={Unifying AI tutor evaluation: An evaluation taxonomy for pedagogical ability assessment of LLM-powered AI tutors},
  author={Maurya, Kaushal Kumar and Srivatsa, Kv Aditya and Petukhova, Kseniia and Kochmar, Ekaterina},
  booktitle={Proceedings of the 2025 Conference of the Nations of the Americas Chapter of the Association for Computational Linguistics: Human Language Technologies (Volume 1: Long Papers)},
  pages={1234--1251},
  year={2025}
}

@inproceedings{macina2025mathtutorbench,
  title={Mathtutorbench: A benchmark for measuring open-ended pedagogical capabilities of llm tutors},
  author={Macina, Jakub and Daheim, Nico and Hakimi, Ido and Kapur, Manu and Gurevych, Iryna and Sachan, Mrinmaya},
  booktitle={Proceedings of the 2025 Conference on Empirical Methods in Natural Language Processing},
  pages={204--221},
  year={2025}
}

@inproceedings{scarlatos2026simulated,
  title={Simulated Students in Tutoring Dialogues: Substance or Illusion?},
  author={Scarlatos, Alexander and Lee, Jaewook and Woodhead, Simon and Lan, Andrew},
  booktitle={Proceedings of the 64th Annual Meeting of the Association for Computational Linguistics (Volume 1: Long Papers)},
  pages={42349--42385},
  year={2026}
}

@article{duan2026history,
  title={Who Am I? History-Aware Profiles for Student Simulation in Tutoring Dialogues},
  author={Duan, Zhangqi and Huang, Shuyan and Scarlatos, Alexander and Lee, Jaewook and Woodhead, Simon and Lan, Andrew},
  journal={arXiv preprint arXiv:2605.30051},
  year={2026}
}

@inproceedings{niousha2026parastudent,
  title={ParaStudent: Closing the Sim2Real Gap in User Simulators for AI Tutor Evaluation},
  author={Niousha, Rose and Miroyan, Mihran and O’Neill, Abigail and Gonzalez, Joseph E and Ranade, Gireeja and DeNero, John and Norouzi, Narges},
  booktitle={Proceedings of the 2026 Conference on Empirical Methods in Natural Language Processing},
  pages={56493--56511},
  year={2026}
}

@inproceedings{chen2024bgem3,
  author = {Chen, Jianlyu and Xiao, Shitao and Zhang, Peitian and Luo, Kun and Lian, Defu and Liu, Zheng},
  title = {{M3-Embedding: Multi-Linguality, Multi-Functionality, Multi-Granularity Text Embeddings Through Self-Knowledge Distillation}},
  booktitle = {Findings of the Association for Computational Linguistics: ACL 2024},
  pages = {2318--2335},
  year = {2024},
  publisher = {Association for Computational Linguistics},
  doi = {10.18653/v1/2024.findings-acl.137},
  url = {https://aclanthology.org/2024.findings-acl.137/},
}

@inproceedings{cormack2009rrf,
  author = {Cormack, Gordon V. and Clarke, Charles L. A. and B{\"u}ttcher, Stefan},
  title = {{Reciprocal rank fusion outperforms condorcet and individual rank learning methods}},
  booktitle = {Proceedings of the 32nd International ACM SIGIR Conference on Research and Development in Information Retrieval},
  pages = {758--759},
  year = {2009},
  publisher = {ACM},
  doi = {10.1145/1571941.1572114},
  url = {https://doi.org/10.1145/1571941.1572114},
}

@misc{eedi2026misconceptions,
  author = {Eedi},
  title = {{Eedi Misconceptions Graph v1.0}},
  year = {2026},
  howpublished = {\url{https://huggingface.co/datasets/Eedi/Eedi-Misconceptions-Graph}},
}

@inproceedings{liu2023xes3g5m,
  author = {Liu, Zitao and Liu, Qiongqiong and Guo, Teng and Chen, Jiahao and Huang, Shuyan and Zhao, Xiangyu and Tang, Jiliang and Luo, Weiqi and Weng, Jian},
  title = {{XES3G5M: A Knowledge Tracing Benchmark Dataset with Auxiliary Information}},
  booktitle = {Advances in Neural Information Processing Systems, Datasets and Benchmarks Track},
  volume = {36},
  pages = {32958--32970},
  year = {2023},
  doi = {10.52202/075280-1429},
  url = {https://papers.nips.cc/paper_files/paper/2023/hash/67fc628f17c2ad53621fb961c6bafcaf-Abstract-Datasets_and_Benchmarks.html},
}

@misc{moe2022compulsory,
  author = {{Ministry of Education of the People's Republic of China}},
  title = {{Compulsory Education Curriculum Standards (2022 Edition)}},
  year = {2022},
  howpublished = {\url{http://www.moe.gov.cn/srcsite/A26/s8001/202204/t20220420_619921.html}},
  note = {Official Ministry of Education notice, 21 April 2022},
}

@article{robertson2009bm25,
  author = {Robertson, Stephen and Zaragoza, Hugo},
  title = {{The Probabilistic Relevance Framework: BM25 and Beyond}},
  journal = {Foundations and Trends in Information Retrieval},
  year = {2009},
  publisher = {Now Publishers},
  doi = {10.1561/1500000019},
  url = {https://doi.org/10.1561/1500000019},
}

@article{team2026kimi,
  title={Kimi k3: Open frontier intelligence},
  author={Team, Kimi and Bai, Tongtong and Bai, Yifan and Bao, Yiping and Cai, Jianfeng and Cai, Xinyuan and Cao, Peizhou and Cao, Yuxuan and Chai, Ziwei and Charles, Y and others},
  journal={arXiv preprint arXiv:2607.24653},
  year={2026}
}

@article{zeng2026glm,
  title={Glm-5: from vibe coding to agentic engineering},
  author={Zeng, Aohan and Lv, Xin and Hou, Zhenyu and Du, Zhengxiao and Zheng, Qinkai and Chen, Bin and Yin, Da and Ge, Chendi and Huang, Chenghua and Xie, Chengxing and others},
  journal={arXiv preprint arXiv:2602.15763},
  year={2026}
}

@techreport{openai2026gpt61sol,
  author      = {{OpenAI}},
  title       = {Addendum to {GPT-6 Astra} System Card: {GPT-6.1 Sol}},
  institution = {OpenAI},
  type        = {System card addendum},
  year        = {2026},
  month       = sep,
  url         = {https://deploymentsafety.openai.com/gpt-6-1-sol},
  urldate     = {2026-11-09},
  note        = {Accessed: 2026-11-09}
}

@techreport{anthropic2026claudeopus55,
  author      = {{Anthropic}},
  title       = {{Claude Opus 5.5} System Card},
  institution = {Anthropic},
  type        = {System card},
  year        = {2026},
  month       = sep,
  url         = {https://www.anthropic.com/claude-opus-5-5-system-card},
  urldate     = {2026-11-09},
  note        = {Accessed: 2026-11-09}
}

@techreport{deepmind2026gemini4argon,
  author      = {{Google DeepMind}},
  title       = {{Gemini 4 Argon} Model Card},
  institution = {Google DeepMind},
  type        = {Model card},
  year        = {2026},
  month       = oct,
  url         = {https://deepmind.google/models/model-cards/gemini-4-argon/},
  urldate     = {2026-11-09},
  note        = {Accessed: 2026-11-09}
}
}

\clearpage
\appendix
\maketitlesupplementary

\providecommand{\na}{---}
\renewcommand{\paragraph}[1]{\par\vspace{.35em}\noindent\textbf{#1}\hspace{0.45em}}
\setcounter{figure}{0}
\setcounter{table}{0}
\setcounter{equation}{0}
\renewcommand{\thefigure}{S\arabic{figure}}
\renewcommand{\thetable}{S\arabic{table}}
\renewcommand{\theequation}{S\arabic{equation}}
\renewcommand{\theHfigure}{S\arabic{figure}}
\renewcommand{\theHtable}{S\arabic{table}}
\renewcommand{\theHequation}{S\arabic{equation}}

\noindent This supplement details the implementation (Section~\ref{sec:s-construction}),
sources and evaluation protocol (Section~\ref{sec:s-annotation}), and additional
resource, grounding and teaching-decision analyses (Sections~\ref{sec:s-e1}--\ref{sec:s-e3}).
Reported counts and results retain the frozen experimental snapshots.

\section{Implementation Details of EduGraph}
\label{sec:s-construction}

This section complements the main paper's three method components: curriculum
infrastructure, evidence grounding and state, and the agent interface. Resource
statistics are collected in Section~\ref{sec:s-e1}; evaluation snapshots and
reproducibility settings are specified in Section~\ref{sec:s-annotation}.

\subsection{Curriculum infrastructure and stable identities}

Models propose Concepts, content and relations; deterministic gates validate the
schema, endpoints, containment forest and prerequisite DAG before release.
Table~\ref{tab:s-pipeline} summarizes the documented process. Stage scripts
01--13 are described in the construction record but are not part of the paper
package; released files support the statistics in Table~\ref{tab:s-release}.

\begin{table*}[t]
\centering
\footnotesize
\setlength{\tabcolsep}{3pt}
\caption{\textbf{Construction stages and deterministic gates.} Proposals enter
the release after structural checks; semantic quality is assessed separately
in Section~\ref{sec:s-e1}.}
\label{tab:s-pipeline}
\begin{tabular*}{\linewidth}{@{\extracolsep{\fill}}>{\raggedright\arraybackslash}p{0.14\textwidth}>{\raggedright\arraybackslash}p{0.40\textwidth}>{\raggedright\arraybackslash}p{0.40\textwidth}@{}}
\toprule
Stage & Operations & Checks \\
\midrule
Requirements & Parse compulsory standards and supplementary senior examination
outlines; retain theme hierarchy and alignable items. & Field whitelist;
separate content requirements from non-alignable rows. \\
Knowledge core & Propose Concepts and path-based keys; merge them into the
\texttt{REFINES} forest. & Unique keys; one containment parent; no cycles or
undeclared key loss. \\
Prerequisites & Propose within-topic and bridge relations; merge duplicates and
reduce transitive edges. & Valid endpoints and stage direction; no cycles or
containment contradictions. \\
Teaching content & Author KeyPoints and Exercises; review with a second model
family. & Field and length limits; provenance and hash keys; option/weight
constraints; hard faults sent to a worklist. \\
Standard alignment & Build the release index; link requirements by explicit
reference and literal name. & Reference score 1.0; literal score below 1;
at most 20 literal links per item; index self-check. \\
Textbook views & Parse official catalogs; map chapters to existing Concepts;
emit an immutable package. & Valid identifiers; missing sources skipped;
parse errors abort; restricted fields removed; recount, leak scan and
idempotent emission. \\
\bottomrule
\end{tabular*}
\end{table*}

A standard item may cover several Concepts, and a Concept can recur in multiple
requirements and textbook views. The grade-3--4 example in main-paper Figure~2
covers decimals, fractions, fraction units, and like-denominator fraction and
one-decimal addition/subtraction through four \texttt{COVERS} edges
(method \texttt{standard-ref}, score 1.0). These Concepts share the path
\texttt{mathematics.primary.\allowbreak number-and-algebra.\allowbreak number-and-operations}
under two sibling Topics. Their 67 \texttt{TEACHES} links span 26 volumes from
six publishers (61 primary, six review). Shared keys and containment paths
stay fixed; grade, volume and unit order belong to each textbook view.
Catalog edits therefore do not create edition-specific Concept identities.

Keys encode subject, stage and up to four English path segments. They are never
reused: renames and merges append aliases, splits append tombstones with
successors, and removals append tombstones without successors. Alias chains
must be acyclic, terminate at an active key, and contain at most eight hops;
undeclared disappearance aborts the build. The registry contains 6,591 active
keys, 4,395 aliases (4,308 renames, 87 merges) and one split tombstone.
Version levels track resolvable identities: patches change content, minor
releases add keys or aliases, and major changes introduce tombstones or change
the key format. Evaluated snapshots are listed in Section~\ref{sec:s-snapshots}.

\subsection{Grounding adapters and learner-state projection}
\label{sec:s-runtime}

The RQ2 benchmark combines lexical and dense retrieval, topic-subtree expansion,
up to two prerequisite hops, role-specific channels and a fitted linear score.
The runtime adapter instead uses BM25 over names, aliases and paths,
a subject/stage prior, an exercise-text index, KeyPoint-derived Skills and
pitfall proxies, and one-hop prerequisites capped at six incoming edges per
seed; binding is deterministic. These adapters share the data and evidence
contract, but the runtime does not execute the strongest RQ2 operator end to end.

For numeric outcomes assigned to Concept $k$, the stored projection uses
\begin{equation}
 w_i=2^{-(t-t_i)/H}c_i,\quad
 \widehat m_k=\frac{\sum_i w_i y_i}{\sum_i w_i},\quad
 q_k=\frac{\sum_i w_i}{\sum_i w_i+2},
\end{equation}
where $y_i$ is an outcome and $c_i$ is binding confidence. The runtime and XES
adapter use $H=30$ days; the dialogue adapter uses a six-turn half-life.
The runtime uses an explicit evaluation time, or otherwise the maximum recorded
time in the evidence prefix. XES uses the last observed attempt time; dialogue
replay uses the current prefix's turn index. The main paper's
$\mathrm{Project}(E_{\le n},R)$ therefore fixes the evaluation time and adapter
settings as well as the evidence prefix and graph.

The runtime accepts an explicit outcome only for an unambiguous Concept target;
it does not propagate grades through prerequisite or containment neighbors.
Ungraded observations add evidence and advance the revision without changing
the graded-outcome summary. The historical RQ3 dialogue adapter additionally
codes binder-labelled correct-answer and understanding turns as 1, incorrect
answers as 0, and other turns, including confusion, as missing. Understanding
includes stated comprehension or a correct restatement: these are model-derived
proxies, not externally supplied grades. The distinction is in evidence
construction; the projector aggregates only non-missing outcomes. The weighted
mean is a heuristic performance summary, not a calibrated mastery posterior.

\subsection{Agent interface and policy scope}

The interface is distributed: \texttt{state}, \texttt{ingest} and
\texttt{decide} belong to the agent object, \texttt{search} to the graph read
model, and the RQ3 position-card \texttt{context} to an evaluation adapter.
They share read-only graph access and append-only evidence rather than one
deployed facade. The deterministic reference policy is separate from the
evaluated LLM policies. Its prerequisite filter requires retained review status
\texttt{reviewed} or \texttt{approved} and confidence at least 0.5, admitting
at most the initial 3,238 edges. This is a metadata filter, not a statement
about teacher confirmation. The action contract admits nine actions, while
the reference policy emits the six evaluated in Section~\ref{sec:s-e3}.

\section{Data Sources and Evaluation Protocol}
\label{sec:s-annotation}
\label{sec:s-repro}

\subsection{Sources and reference-label lineage}

Curriculum construction uses China's 2022 compulsory-education standards, with
senior examination outlines as a supplementary alignment source rather than the
complete senior standards corpus. Table~\ref{tab:s-sources} identifies the
external resources and their experimental roles. Group-level splits and
task-specific sampling are detailed in Sections~\ref{sec:s-e2} and~\ref{sec:s-e3}.

\begin{table*}[t]
\centering
\footnotesize
\setlength{\tabcolsep}{3pt}
\caption{\textbf{External sources and their roles.} Route nodes and leaf
knowledge components are distinct units; private source text is not released.}
\label{tab:s-sources}
\begin{tabular*}{\linewidth}{@{\extracolsep{\fill}}>{\raggedright\arraybackslash}p{0.18\textwidth}>{\raggedright\arraybackslash}p{0.45\textwidth}>{\raggedright\arraybackslash}p{0.30\textwidth}@{}}
\toprule
Source & Representation used & Role \\
\midrule
XES3G5M~\cite{liu2023xes3g5m} & 7,652 questions, 865 leaf KCs;
1,175 KC-route nodes in our alignment representation;
7,295,810 raw-file interactions. & RQ1 alignment; RQ2 wrong-answer evidence;
RQ3 trajectories. \\
K12-KGraph~\cite{liang2026k12} & 10,685 nodes and 23,278 edges, including
6,574 Concepts, 2,518 prerequisite edges, 1,174 Exercises and 48 books. &
Resource comparison and RQ1 bidirectional coverage. \\
Eedi~\cite{eedi2026misconceptions} & 2,258 constructs and 8,117 misconceptions. &
Resource comparison only. \\
Private homework and dialogue & Homework alignment: 840 rows, 535 distinct
labels; observation and dialogue subsets specified per benchmark. &
RQ1 label alignment; RQ2 grounding; RQ3 dialogue decisions. \\
\bottomrule
\end{tabular*}
\end{table*}

Final consolidated labels were model-prefilled and confirmed by two teachers.
RQ1 quality prefill used GPT-6.1-Sol, Claude-Opus-5.5 and Gemini-4-Argon;
alignment used a primary judge and an independent-family second judge on a
sample. RQ2 and RQ3 used those three models plus Kimi-K3 and GLM-5.3.
Reported $\kappa$ values measure agreement among prefilling models before
confirmation, not teacher--teacher agreement. The documented process establishes
teacher confirmation; it does not supply a separate inter-teacher agreement
estimate or independent adjudication statistic.

\paragraph{Evaluation references.}
Consolidated labels and the references actually used for scoring must be
distinguished. Evaluated panel members use stored leave-one-out (LOO) references;
non-panel backbones use the full panel. RQ2 uses entity-level majority rules,
whereas RQ3 uses plurality with a confidence tie-break. The RQ3 audit found that
some LOO labels differ from the final teacher-confirmed labels; the reported
accuracies were not rescored. Accordingly, agreement is with the stored
evaluation reference and does not imply that every model-specific LOO label
was independently teacher-confirmed. Sections~\ref{sec:s-e2} and~\ref{sec:s-e3}
give the reference-specific denominators, missing-vote handling and metrics.

\subsection{Evaluation snapshots and statistical settings}
\label{sec:s-snapshots}

RQ1 counts use EduGraph's fixed evaluation snapshot. Its 4,882 prerequisites
comprise 3,238 initial edges and 1,644 added edges; 742 transitive-reduction
rows are excluded. The added edges do not change Concept keys.
Retained \texttt{candidate} metadata is not a teacher-annotation
status. RQ2 retrieval and the EduGraph binder use the unified layer, while stored
binder baselines and ablations use the initial 3,238-edge annotation; their
difference is not a controlled typed-view ablation. RQ3 uses a separately
staged EduGraph snapshot combining the base graph with an evaluation overlay.

Sampling uses seed 20261105 for quality, alignment and K12-KGraph samples.
RQ2/RQ3 model runs use seeds 0/1/2 at temperature 0.3, except the explicitly
listed incomplete baseline runs. RQ2 retrieval intervals use 1,000
group-bootstrap resamples (seed 0); RQ3 uses 2,000. The headline prerequisite
quality interval uses effective-sample-size Wilson. A 100,000-draw stratified
parametric bootstrap (seed 20261105) and finite-population-corrected Wilson
are sensitivity analyses, yielding 90.5--95.7\% and 90.2--95.3\%, respectively;
the sampling-frame restriction is discussed in Section~\ref{sec:s-e1}.

\subsection{Artifacts, data handling and reproducibility}

The paper package contains manuscript and figure sources, sanitized aggregates,
validation scripts, and the four rewritten English cases in main-paper
Figures~5--6, with selected stored predictions and teacher-confirmed references.
The graph, schema, runtime and evaluation code are separate research resources.
Full benchmark text, original identifiers, model outputs and row-level labels
remain private; exact reruns additionally require the private sources and
historical model endpoints. The included validator checks selected quantities
against aggregates; separate experiment resources provide \texttt{run\_all.sh}
entry points. These availability levels should not be read as a claim that the
paper package alone reproduces every model call.

Textbook releases retain publisher, series, edition, grade, volume and catalog
structure (chapter identifier, parent, level, order and available page starts),
but omit chapter titles and printed numbers. Schema, writer and leak checks
enforce this boundary. RQ1's \texttt{TEACHES} audit used private titles and
publishes only aggregate verdicts. Case grouping uses pseudonymous identifiers;
no student names or contact details are released. No new student data were
collected and no student behavior was experimentally changed.

Fixed proposal files and gate code yield byte-identical releases. Model requests
are cached by the complete request hash; a cache hit reuses the stored proposal,
rather than demonstrating deterministic fresh model generation. The release
verifier reports zero validation errors and 7,329 non-blocking metadata warnings.
Dense retrieval and reranking run on CPU; model calls use a shared cache and
cost ledger. Endpoints and credentials remain in environment variables.
Per-query costs are reported in Section~\ref{sec:s-e2}; aggregate spending is
not separated from development and extension calls in the ledger.

\section{Additional Resource Analyses (RQ1)}
\label{sec:s-e1}

This section supplements the main paper's resource landscape, textbook views,
coverage and precision analyses. Counts are recomputed from the frozen release;
annotation conventions are given in Section~\ref{sec:s-annotation}.

\subsection{Resource composition and graph structure}

Table~\ref{tab:s-release} gives the subject and stage breakdown underlying the
main paper's Figure~3a. The knowledge core additionally contains 1,322 Topics
and 6,417 REFINES edges. Auxiliary resources comprise 2,844 standard/outline
items (1,660 alignable), 7,694 COVERS edges, 6,878 textbook chapters and 21,971
TEACHES edges. KeyPoints comprise 5,268 definitions, 2,628 properties, 2,454
methods, 590 formulas, 426 connections and 399 pitfalls; TEACHES roles are
19,985 primary, 1,820 mentioned and 166 review.
By primary/junior/senior stage, KeyPoints number 3,058/5,070/3,637;
Exercises 2,217/3,171/1,719; outgoing prerequisites 1,495/2,182/1,205; and
volumes 344/266/142.

The standards corpus comprises 1,956 compulsory-education items and 888
senior examination-outline items. Its 1,660 alignable items comprise 772
compulsory requirements and all 888 outline items; the remaining 1,184
compulsory rows are non-alignable. Concept resources occupy 24 subject--stage
partitions; \texttt{english/unspecified} and \texttt{politics/unspecified}
contain standards only.

\begin{table*}[t]
\centering
\footnotesize
\setlength{\tabcolsep}{2.0pt}
\caption{\textbf{EduGraph by subject and school stage.}
Concepts are split by stage; other columns give subject totals. Prerequisite
edges are counted once by source Concept; none crosses subjects. Separately
catalogued edition variants count as distinct volumes.}
\label{tab:s-release}
\begin{tabular*}{\linewidth}{@{\extracolsep{\fill}}lrrrrrrrr@{}}
\toprule
 & \multicolumn{4}{c}{Concepts} & & & & \\
\cmidrule(lr){2-5}
Subject & primary & junior & senior & all & KeyPoints & Exercises & Prereq.\ from & Volumes \\
\midrule
Science       & 536 & 594 & 0   & 1,130 & 2,765 & 2,194 & 1,012 & 94 \\
English       & 100 & 253 & 340 & 693   & 1,423 & 693   & 675   & 250 \\
Politics      & 290 & 103 & 269 & 662   & 1,495 & 994   & 447   & 44 \\
Mathematics   & 166 & 143 & 323 & 632   & 1,440 & 669   & 876   & 184 \\
Chinese       & 236 & 133 & 65  & 434   & 941   & 570   & 300   & 41 \\
Biology       & 0   & 301 & 104 & 405   & 892   & 559   & 361   & 35 \\
History       & 0   & 281 & 124 & 405   & 866   & 528   & 220   & 17 \\
Chemistry     & 0   & 150 & 238 & 388   & 844   & 388   & 463   & 24 \\
Physics       & 0   & 223 & 142 & 365   & 771   & 357   & 397   & 28 \\
Geography     & 0   & 77  & 78  & 155   & 328   & 155   & 131   & 35 \\
\midrule
Total         & 1,328 & 2,258 & 1,683 & 5,269 & 11,765 & 7,107 & 4,882 & 752 \\
\bottomrule
\end{tabular*}
\end{table*}

The PREREQUISITE DAG reaches 79.8\% of Concepts; its longest path has 11 edges
and 12 Concepts across all three stages of mathematics. Stage-pair counts are primary to
primary/junior/senior: 672/676/147; junior to junior/senior: 1,336/846; senior
to senior: 1,205. Thus 1,669 edges cross stages, as summarized in the main
paper's Figure~3b. Mean in- and out-degree are both 0.93 (maxima 16 and 26),
with 2,099 sources and 2,989 sinks. Densification reduces isolated Concepts
from 31.9\% to 20.2\% (1,065), leaving 1,341 weakly connected components,
the largest with 569 Concepts. Isolation ranges from 2.7\% in mathematics
(17/632) to 39.5\% in history (chemistry 6.2\%, physics 13.2\%); absent edges
do not establish that no prerequisite exists.

Internal association coverage is 5,259/5,269 (99.81\%) for Concepts having
both a KeyPoint and an Exercise, 4,922/5,269 (93.41\%) for Concepts with a
COVERS link, and 4,918/5,269 (93.34\%) for Concepts citing a standard item in
\texttt{standard\_refs}. In the reverse direction, 1,514/1,660 alignable
standard/outline items have COVERS links (91.2\%). These measure association
presence; mapping precision is audited separately below.

\subsection{Textbook views and ordering evidence}

A \emph{volume} is a distinct catalogued book; a \emph{line} is a
(publisher, subject) curriculum sequence joining its volumes across stages and
series. The catalogue contains 752 volumes, 20 publishers and 54
publisher--subject pairs. Mapped resources comprise 527 volumes (70.1\%),
3,138/6,878 chapters (45.6\%), and 3,789/5,269 Concepts (71.9\%) with any
TEACHES role. Only primary-role anchors define first teaching positions:
3,416 Concepts (64.8\%) across 47 lines from 18 publishers, yielding 10,753
Concept--line pairs. The other two publishers have no mapped chapter. All 225
unmapped volumes have unit-only contents metadata
(\texttt{toc\_completeness=unit\_only\_gap}); 321 additional unit-only volumes
nevertheless have an anchor. Missing mappings do not imply that a book omits
the knowledge.

Publisher views preserve organizational differences. Among 161,124 eligible
Concept pairs co-taught in the same grade, 15.9\% appear in opposite orders
across publishers. Rates are below 1\% in biology, chemistry, physics and
geography, 21.5\% in science and 14.7\% in mathematics; English's 63.8\% rests
on only 69 pairs. Median Kendall $\tau_b$ is .490, with 49.3\% of cell
estimates at least .5. Of 2,569 multi-publisher Concepts, 755 (29.4\%) have
different first teaching grades and 488 (19.0\%) differ by exactly one grade.
A Concept is taught by 3.15 publishers on average (5.43 in mathematics).
These statistics extend the main paper's Figure~2c using teaching positions;
chapter titles are not released.

Textbook order checks prerequisite direction using a source separate from
the prerequisite-proposal stage. For 1,890 observations with same-stage
endpoints taught in different units of one line, order agrees in 78.3\% of
cases (95\% CI 75.0--81.6), versus 50.7\% for matched random pairs
(49.5--51.9) and 50.0\% for REFINES siblings (47.4--52.7). Agreement is
88.7\% in mathematics (84.9--91.9) but 51.0\% in integrated science
(42.1--59.9). The aggregate supports an ordering signal, with substantial
subject variation.

\subsection{External alignment}

BM25 over Concept names, aliases, paths and KeyPoint titles retrieves 20
candidates; CPU bge-m3~\cite{chen2024bgem3} retrieves another 20.
Reciprocal-rank fusion ($k=60$) retains 12~\cite{cormack2009rrf}. Given names,
paths, stages and up to five KeyPoint titles, the judge assigns exact, broader,
narrower, related or none; coverage counts exact or broader matches. The
primary judge is GPT-6.1-Sol. Claude-Opus-5.5 repeats a random 300-item XES
sample with the same candidates and prompt (299 usable): agreement is 80.3\%,
with Cohen's $\kappa=.722$ for five labels and $.759$ for covered/not covered.
Its coverage is 51.17\% (95\% CI 45.53--56.79), versus the primary judge's
55.57\%; the 4.4-point difference indicates judge sensitivity.

For photographed homework, deduplicating 840 rows yields 535 distinct labels
and 76.1\% coverage (95\% CI 72.3--79.5), versus 77.1\% when row-weighted;
the latter interval does not account for repeated labels. Row-level coverage
ranges from biology's 39.4\%, where labels describe finer-grained processes,
to chemistry's 89.3\%; Chinese, physics, mathematics, geography and English
reach 81.3\%, 81.2\%, 79.6\%, 70.0\% and 66.3\%, respectively. History has
only one row and is not interpreted.

XES alignment is stronger for curriculum content than extension content.
Module rates are 78.45\% (116 in-curriculum Concepts), 74.55\% (224
Concepts), 51.72\% (58 in-curriculum problem types), 48.21\% (755 extension
Concepts), 25.00\% (four mathematical-idea Concepts) and 0\% (18 ability and
thinking Concepts); 17.2\% of all components are unrelated. K12-KGraph
alignment is direction-dependent: mapping its Concepts into EduGraph yields
16.0\% exact and 52.75\% broader matches, whereas the reverse direction gives
48.75\% exact matches. Both directions cover roughly two-thirds of the sampled
Concepts, reflecting different granularity.

These resources support different tasks: K12-KGraph has 10,685 nodes and
23,278 edges, including 6,574 Concepts, 2,518 Concept prerequisites,
1,174 Exercises and 48 books across four subjects;
XES provides 7,652 questions and 865 leaf KCs (1,175 KC-route nodes in our
representation), whose hierarchy is not an explicit prerequisite graph; the
Eedi resource contains 2,258 constructs and 8,117 misconceptions. EduGraph
additionally connects standards, textbook views and learner evidence through
versioned Concept identities. Source uses and release boundaries are recorded
in Section~\ref{sec:s-annotation}.

\subsection{Quality audit and sensitivity}

The audit independently samples 200 definitions from 5,268, 200 Exercises
from 7,107, 200 COVERS edges from 7,694 and 200 TEACHES edges from 21,971.
Sampling is stratified by subject, and also by stage for definitions and
Exercises (seed 20261105). PREREQUISITE uses two 200-edge samples: one from
the 3,238-edge initial stratum and one from the 2,386-row pre-reduction overlay;
122 rows of the latter survive in the final 1,644-edge added stratum.
Three-model prefills use majority \emph{correct/incorrect/unsure} verdicts,
with ties marked \emph{unsure}, followed by teacher confirmation
(Section~\ref{sec:s-annotation}). Decided precision is
correct/(correct+incorrect); conservative precision counts \emph{unsure} as
not correct. Layers have different sampling units and are not pooled into an
overall precision.

\begin{table*}[t]
\centering
\footnotesize
\setlength{\tabcolsep}{3.2pt}
\caption{\textbf{Quality audit by graph layer.} \emph{Cons.} counts unsure as
not correct; $\kappa$ is prefill-model Fleiss agreement, not teacher agreement.
Intervals are Wilson 95\%, with an effective-$n$ interval for the combined
PREREQUISITE estimate. Its overlay sample predates reduction; only 122 rows
remain in the final view. TEACHES was judged using private chapter titles.}
\label{tab:s-quality}
\begin{tabular*}{\linewidth}{@{\extracolsep{\fill}}lrrrrcccc@{}}
\toprule
Category & Judged & Correct & Incorrect & Unsure & Precision (\%) & 95\% CI & Cons. (\%) & $\kappa$ \\
\midrule
Concept definition & 200 & 197 & 3 & 0 & 98.5 & 95.7--99.5 & 98.5 & 0.288 \\
Exercise (answer + explanation) & 200 & 195 & 3 & 2 & 98.5 & 95.6--99.5 & 97.5 & 0.648 \\
COVERS edge & 200 & 119 & 66 & 15 & 64.3 & 57.2--70.9 & 59.5 & 0.452 \\
TEACHES edge & 200 & 122 & 21 & 57 & 85.3 & 78.6--90.2 & 61.0 & 0.393 \\
\midrule
PREREQUISITE, release-stratum sample & 200 & 186 & 14 & 0 & 93.0 & 88.6--95.8 & 93.0 & 0.381 \\
PREREQUISITE, added-edge overlay sample & 200 & 187 & 13 & 0 & 93.5 & 89.2--96.2 & 93.5 & 0.667 \\
\textbf{PREREQUISITE layer (4,882)} & \textbf{400} & \textbf{373} & \textbf{27} & \textbf{0} & \textbf{93.2} & \textbf{90.1--95.3} & \textbf{93.2} & -- \\
\bottomrule
\end{tabular*}
\end{table*}

Table~\ref{tab:s-quality} supplies the counts behind the main paper's
Figure~3d. COVERS illustrates why presence is insufficient: 91.2\% item
coverage coexists with 64.3\% decided edge precision. English reaches 19.4\%
(6/31 decided edges), often linking broad ability requirements to Concepts
representing only part of the requirement; geography reaches 16.7\%.
Other subject rates are Chinese 61.5\%, physics/chemistry 66.7\%, mathematics
70.0\%, politics 76.2\%, biology/history 76.9\% and science 87.5\%.
Prefill correct rates vary most for COVERS (50.5/77.0/50.5\%) and least for
definitions (99.5/98.0/97.5\%); the latter's low $\kappa$ reflects a ceiling
effect. TEACHES has 28.5\% unsure cases, explaining its 85.3\% decided versus
61.0\% conservative precision.

The PREREQUISITE estimate weights initial and added strata by
3,238/4,882 and 1,644/4,882. The stored 93.2\% uses the full overlay sample's
93.5\% as a proxy for the added stratum, transferring a broader sampling
frame to the final population. Its 90.1--95.3\% interval uses effective-$n$
Wilson ($n_{\mathrm{eff}}=358.3$), not 400 identically sampled edges; finite
population correction gives 90.2--95.3\%. Restricting to the 122 sampled rows
in the final view gives 114/122 $=93.4\%$ (Wilson 87.6--96.6). Neither interval
eliminates the sampling-frame mismatch.

\section{Additional Grounding Results (RQ2)}
\label{sec:s-e2}

This section supplies the benchmark protocol, complete retrieval results,
binding results, and ablations behind RQ2. Question targets describe what an
item assesses; learner issues describe difficulties supported by an attempt.
Retrieval and role binding are evaluated separately.

\subsection{Benchmark and protocol}

The 1,000 real K12 observations are divided into 200 development and 800 test
cases at the student, session, or item-family level (956 groups), with no group
crossing splits. Test/development counts are 332/83 photographed homework
questions, 144/36 wrong answers with prior attempts, 105/26 follow-ups,
155/39 tutoring-dialogue turns, and 64/16 feedback cases. Homework is stratified
by subject and stage within covered partitions; primary physics, chemistry,
biology, history, and geography are absent. XES cases require a name-level
correspondence to an EduGraph mathematics Concept so their targets can be annotated.

Candidate pools combine subject/stage-restricted BM25~\cite{robertson2009bm25},
CPU bge-small-zh-v1.5 retrieval, the agent's lexical search, and a non-panel
model's proposals matched to graph entities, including up to six cross-subject
slots. Expansion adds skills, misconception proxies, prerequisites, and
problem-type cues. Each pool contains 36--57 nodes (mean 48.6, median 49) and
is shuffled to hide retrieval rank.

The five-model panel in Section~\ref{sec:s-annotation} labels this same pool.
Targets admit Concept, Skill, and Problem Type; issues admit Concept, Skill,
Prerequisite, and Misconception. Multi-selection is allowed, with explicit
\emph{none} and \emph{unclear} states. A wrong answer alone does not establish
a misconception; no student work requires \texttt{learner\_issue=none}.
An entity needs a strict majority of available votes (three of five); status is
resolved separately if the majority set is empty. Evaluated panel members use
the other four models' leave-one-out reference; non-panel backbones use the
full panel. These are the stored evaluation references; teacher confirmation
and reference provenance are distinguished in Section~\ref{sec:s-annotation}.
Prefill Fleiss $\kappa$ is .577/.479 and mean pairwise micro-F1 .683/.410 for
targets/issues, respectively, not teacher agreement. Under the full-panel
reference, test target statuses are 712/71/17 present/none/unclear and issue
statuses are 137/350/313.

\begin{figure*}[t]
\centering
\includegraphics[width=\textwidth]{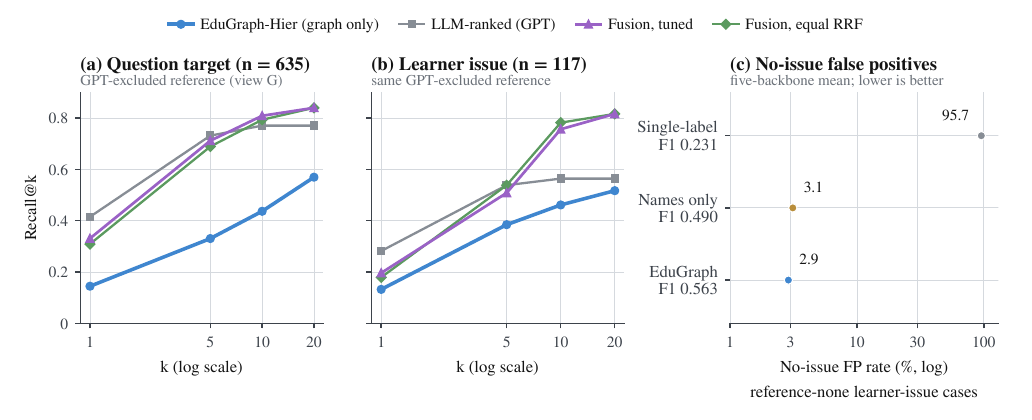}
\caption{\textbf{Retrieval and binding have different failure modes.}
(a, b) Recall@$k$ under the GPT-excluded reference, with the same 635 target
or 117 issue cases for every curve; $k$ uses a log scale. Tuned fusion is best
at $k=10$ for targets, equal-weight fusion for issues. (c) Binding on the fixed
pool, averaged over five backbones: reference-none issue false-positive rate
(log scale, lower is better), with all-case issue F1 in the labels.
The EduGraph annotation uses 4,882 edges and stored baselines use 3,238.}
\label{fig:s1-grounding}
\end{figure*}

\paragraph{Task A: catalog retrieval.}
Search covers all 5,269 Concepts, not the annotated pool. Global baselines score
all nodes; routed methods search the subject/stage partition and supplement it
with the outside global ranking at half weight. Skill and Misconception
key-point IDs project to their parent Concept; prerequisite proxies retain
Concept IDs; Problem Type labels lack a Concept target and are excluded.
The full-panel reference yields 673 of 712 present target cases and all 137
present issue cases; the GPT-excluded reference yields 635 of 681 and all 117,
respectively. Metrics are mean per-case Recall@$\{1,5,10,20\}$ and MRR, with
group-bootstrap 95\% intervals. Table~\ref{tab:s-e2-retrieval} keeps these
references in separate panels; Figure~\ref{fig:s1-grounding} uses only the
GPT-excluded reference (view G). The pool-level graph-gap flag instead uses
all 800 test cases: 152/800 (19.0\%) under the full panel and 134/800 (16.75\%)
under GPT exclusion. Its complements, .810 and .833, are not catalog-recall
ceilings because the denominators and search spaces differ.

\paragraph{Task B: candidate binding.}
Binding uses the fixed pool and entity IDs without Concept projection. We report
micro-F1 and exact-set match on present cases, all-case micro-F1 on present plus
none cases, false-positive rate on none cases, and abstention on unclear cases.
None contributes an empty reference; unclear is excluded from F1 because it has
no resolved ground truth. Test runs use seeds 0/1/2 at temperature .3 and
development uses seed 0; baseline exceptions are listed with the binding table.

\subsection{Retrieval results}

\begin{table*}[t]
\centering
\footnotesize
\setlength{\tabcolsep}{2.4pt}
\caption{\textbf{Task A: Concept-level retrieval.} Mean per-case Recall@$k$
and MRR over the full 5,269-Concept catalog. Panel A uses the stored five-model
reference (673 question-target and 137 learner-issue cases); Panel B uses the
GPT-excluded reference (635 and 117). References and denominators must not be
mixed across panels. LLM-direct returns at most three labels. EduGraph rows
use the 4,882-edge graph.}
\label{tab:s-e2-retrieval}
\begin{tabular*}{\linewidth}{>{\raggedright\arraybackslash}p{\dimexpr0.26\linewidth-2\tabcolsep\relax}>{\centering\arraybackslash}p{\dimexpr0.074\linewidth-2\tabcolsep\relax}>{\centering\arraybackslash}p{\dimexpr0.074\linewidth-2\tabcolsep\relax}>{\centering\arraybackslash}p{\dimexpr0.074\linewidth-2\tabcolsep\relax}>{\centering\arraybackslash}p{\dimexpr0.074\linewidth-2\tabcolsep\relax}>{\centering\arraybackslash}p{\dimexpr0.074\linewidth-2\tabcolsep\relax}>{\centering\arraybackslash}p{\dimexpr0.074\linewidth-2\tabcolsep\relax}>{\centering\arraybackslash}p{\dimexpr0.074\linewidth-2\tabcolsep\relax}>{\centering\arraybackslash}p{\dimexpr0.074\linewidth-2\tabcolsep\relax}>{\centering\arraybackslash}p{\dimexpr0.074\linewidth-2\tabcolsep\relax}>{\centering\arraybackslash}p{\dimexpr0.074\linewidth-2\tabcolsep\relax}}
\toprule
 & \multicolumn{5}{c}{Question target} & \multicolumn{5}{c}{Learner issue} \\
\cmidrule(lr){2-6}\cmidrule(lr){7-11}
Method & R@1 & R@5 & R@10 & R@20 & MRR & R@1 & R@5 & R@10 & R@20 & MRR \\
\midrule
\multicolumn{11}{l}{\emph{A. Graph-native methods (five-model reference)}} \\
BM25, flat & 0.084 & 0.180 & 0.224 & 0.278 & 0.186 & 0.101 & 0.344 & 0.394 & 0.470 & 0.238 \\
Dense bge-m3, flat & 0.084 & 0.223 & 0.290 & 0.370 & 0.212 & 0.079 & 0.308 & 0.365 & 0.422 & 0.210 \\
Hybrid RRF, flat & 0.103 & 0.235 & 0.292 & 0.380 & 0.236 & 0.130 & 0.358 & 0.419 & 0.484 & 0.276 \\
BM25, routed & 0.116 & 0.266 & 0.331 & 0.391 & 0.263 & 0.080 & 0.172 & 0.223 & 0.304 & 0.153 \\
Dense, routed & 0.128 & 0.286 & 0.364 & 0.474 & 0.297 & 0.062 & 0.186 & 0.218 & 0.302 & 0.150 \\
Hybrid RRF, routed & 0.144 & 0.295 & 0.391 & 0.518 & 0.317 & 0.066 & 0.157 & 0.263 & 0.425 & 0.164 \\
Hybrid + reranker & 0.101 & 0.287 & 0.394 & 0.518 & 0.266 & 0.090 & 0.253 & 0.341 & 0.425 & 0.192 \\
\rowcolor{bestblue} EduGraph-Hier & 0.138 & 0.322 & 0.427 & 0.555 & 0.321 & 0.150 & 0.381 & 0.472 & 0.520 & 0.290 \\
\midrule
\multicolumn{11}{l}{\emph{B. GPT-6.1-Sol operators (leave-one-out GPT reference)}} \\
Hybrid + reranker & 0.109 & 0.301 & 0.404 & 0.519 & 0.256 & 0.080 & 0.284 & 0.373 & 0.450 & 0.195 \\
\rowcolor{bestblue} EduGraph-Hier & 0.145 & 0.331 & 0.437 & 0.570 & 0.306 & 0.133 & 0.385 & 0.462 & 0.517 & 0.268 \\
LLM-direct, GPT & 0.415 & 0.608 & 0.608 & 0.608 & 0.623 & 0.256 & 0.436 & 0.436 & 0.436 & 0.365 \\
LLM-ranked top-10, GPT & 0.415 & 0.732 & 0.771 & 0.771 & 0.659 & 0.282 & 0.539 & 0.564 & 0.564 & 0.423 \\
\rowcolor{bestblue} Hier + LLM, GPT tuned & 0.331 & 0.712 & \textbf{0.809} & \textbf{0.840} & 0.592 & 0.197 & 0.509 & 0.756 & \textbf{0.816} & 0.388 \\
\rowcolor{bestblue} Hier + LLM, GPT equal RRF & 0.309 & 0.689 & 0.793 & \textbf{0.840} & 0.568 & 0.180 & 0.539 & \textbf{0.782} & \textbf{0.816} & 0.379 \\
\bottomrule
\end{tabular*}
\end{table*}

\paragraph{Complementary recall.}
Under the full-panel reference, Hier improves R@10 over the strongest routed
baseline by .033 for targets (95\% CI [.001, .065]) and .131 for issues
([.040, .220]); its issue gain over the strongest flat baseline is .054
([.009, .101]). Under GPT exclusion, however, Hier trails LLM-direct on targets
by .171 ([$-.218$, $-.128$]) and has no resolved advantage on issues
(+.026 [$-.115$, .170], $p=.73$). Fusion combines their rankings: tuned GPT
fusion gains .038 [.025, .053] target R@10 and .192 [.123, .271] issue R@10
over the LLM ranker; equal weights gain .023 [.004, .041] and .218
[.137, .313]. This improves candidate recall, but lowers target R@1
(.415 to .331) and MRR (.659 to .592) for tuned fusion.

For non-panel DeepSeek-V4.1-Flash, tuned fusion reaches .765/.702 target/issue
R@10 under the full-panel reference, improving over its ranker by .023
[.013, .034]/.223 [.142, .302]. Equal weighting has no resolved target gain
(+.002 [$-.014$, .017], $p=.842$), with the same issue gain. Thus neither
weighting is uniformly best. The strongest source-specific issue gain occurs
on wrong-answer histories: GPT ranker, Hier, tuned fusion, and equal fusion
reach .510, .621, .774, and .853 R@10. The .782 aggregate issue recall is a
mean over 117 localized issue cases, not a success rate over all observations;
its any-hit rate is 93/117, versus 67/117 for the ranker.

\paragraph{Ablations and scope.}
Removing the topic layer or routing prior reduces target R@10 by .053
[.035, .071] or .057 [.039, .075]; removing exercise-dense or lexical channels
reduces issue R@10 by .033 [.007, .065] or .029 [.007, .057]. Prerequisite
expansion contributes .015 [0, .039]. Misconception proxies show no measurable
effect with only four labelled cases, and the key-point dense channel is
inactive in this build (lexical key-point evidence remains).
Densification from 3,238 to 4,882 edges leaves issue R@10 at .472 and reduces
MRR by .019--.027 ($p=.008$--.026); the tuned-fusion gain of .0085 is not
significant. The unified graph is retained for coverage, not a measured
retrieval gain. Although Task A searches the full catalog, its references were
annotated from a proposed pool, favoring proposal channels, particularly
DeepSeek, which also proposed candidates. GPT exclusion reduces panel-membership
bias but does not remove this pool-selection bias.

\subsection{Role binding}

\begin{table*}[t]
\centering
\footnotesize
\setlength{\tabcolsep}{3.0pt}
\caption{\textbf{Task B: Binding results on the fixed candidate pool.}
QT F1 uses present question targets; LI F1 present and LI exact use present
learner issues; LI F1 all includes present and none, excluding unclear;
LI FP is the issue rate on reference-none cases. Baseline means cover five
backbones; Qwen3.8-Max lacked a complete grid. Runs use seeds 0/1/2 except the
Kimi-K3 baseline (seed 0 only). Evaluated panel members use their leave-one-out
reference; other backbones use the full panel. The EduGraph candidate annotation
uses 4,882 edges, whereas stored baselines use 3,238; this is not a controlled
ablation of the typed view.}
\label{tab:s-e2-binder}
\begin{tabular*}{\linewidth}{>{\raggedright\arraybackslash}p{\dimexpr0.23\linewidth-2\tabcolsep\relax}>{\raggedright\arraybackslash}p{\dimexpr0.21\linewidth-2\tabcolsep\relax}>{\centering\arraybackslash}p{\dimexpr0.112\linewidth-2\tabcolsep\relax}>{\centering\arraybackslash}p{\dimexpr0.112\linewidth-2\tabcolsep\relax}>{\centering\arraybackslash}p{\dimexpr0.112\linewidth-2\tabcolsep\relax}>{\centering\arraybackslash}p{\dimexpr0.112\linewidth-2\tabcolsep\relax}>{\centering\arraybackslash}p{\dimexpr0.112\linewidth-2\tabcolsep\relax}}
\toprule
Binder & Backbone & QT F1 & LI F1 present & LI F1 all & LI exact & LI FP \\
\midrule
Top-1 lexical & mean & 0.104 & 0.030 & 0.012 & 0.016 & 1.000 \\
Single-label LLM & mean & 0.515 & 0.560 & 0.231 & \textbf{0.450} & 0.957 \\
Role-separated, no graph & mean & 0.593 & 0.523 & 0.490 & 0.251 & 0.031 \\
\midrule
\rowcolor{bestblue} EduGraph binder & GPT-6.1-Sol & 0.752 & 0.483 & 0.462 & 0.311 & 0.032 \\
\rowcolor{bestblue} EduGraph binder & Claude-Opus-5.5 & 0.742 & 0.659 & 0.604 & 0.243 & 0.032 \\
\rowcolor{bestblue} EduGraph binder & Gemini-4-Argon & 0.711 & 0.580 & 0.522 & 0.141 & 0.046 \\
\rowcolor{bestblue} EduGraph binder & DeepSeek-V4.1-Flash & 0.754 & 0.592 & 0.587 & 0.316 & 0.007 \\
\rowcolor{bestblue} EduGraph binder & Kimi-K3 & 0.760 & 0.675 & 0.641 & 0.391 & 0.028 \\
\cmidrule(lr){2-7}
\rowcolor{bestblue} EduGraph binder & mean & \textbf{0.744} & \textbf{0.598} & \textbf{0.563} & 0.280 & \textbf{0.029} \\
\bottomrule
\end{tabular*}
\end{table*}

\paragraph{Role separation and abstention.}
Table~\ref{tab:s-e2-binder} reports baseline means and per-backbone EduGraph
results. The single-label model remains competitive on present issues but emits
an issue on 95.7\% of reference-none cases. Role separation lowers this to
3.1\% with names only; the typed binder reaches 2.9\%. On unclear cases it
abstains 63.7--88.1\% of the time (mean 77.7\%), versus 5.3--18.4\% for the
single-label model. The typed binder's observed gains over names only are .151
in target F1 and .073 in all-case issue F1, but graph annotations differ
(4,882 versus 3,238 edges), so these do not isolate a typed-view effect.
Per-case issue gains range from .013 to .048, significant on four of five
backbones; Kimi-K3 is not significant and its baseline has only one seed.

Performance depends on the evidence source. The binder improves issue F1 on
dialogues (.68 versus .48) and follow-ups (.50 versus .38), but is worse on
wrong-answer histories (.45 versus .54). On present issues, the single-label
model also has higher exact-set match (.450 versus .280). These differences
prevent interpreting the aggregate gain as uniformly better diagnosis.

\paragraph{Controlled binding ablations.}
The archived ablations use the initial 3,238-edge annotation throughout and
were not rerun on the unified graph. For GPT-6.1-Sol and DeepSeek-V4.1-Flash,
respectively, full QT/all-case LI F1 is .757/.475 and .747/.568;
removing prerequisite annotation gives .754/.509 and .748/.603;
removing misconception candidates gives .762/.502 and .742/.583.
These two removals change F1 by less than .04 ($p=.22$--.97).
Removing abstention instead lowers all-case LI F1 to .238 and .311;
stored per-case present-F1 differences are $-.78$ and $-.75$ (both $p<.001$),
and target differences are $-.10$ and $-.05$.
This supports retaining abstention; it does not establish an independent
benefit from the prerequisite or misconception annotations, nor isolate the
typed-view contribution in the unmatched headline comparison.

\subsection{Cost and latency}

Hier runs on CPU without an API call: warm retrieval p50 is 75 ms, bge-m3
encoding 181 ms, and end-to-end p50 256 ms (mean 672 ms). LLM ranking costs
approximately \$0.00547 per GPT-6.1-Sol query and \$0.00149 per
DeepSeek-V4.1-Flash query, with p50 latency 6.6 s and 9.1 s. Fusion adds the
CPU graph path without another API call; its measured benefit is complementary
recall rather than lower end-to-end latency.

\section{Additional Teaching-Decision Results (RQ3)}
\label{sec:s-e3}

RQ3 measures action agreement under fixed observations and stored evaluation
references. It compares direct use of structured context with augmentation of
the same raw history; it does not measure the effect of executing an action.

\subsection{Splits, inputs, and actions}

An earlier 600-case design sample (300 trajectories and 300 dialogue scenes;
100 prompt-development and 500 internal-test cases) informed prompt design and
is not a confirmation split. The \emph{Development Set} contains 724 decision
points (500 trajectories and 224 dialogue turns) inspected before the round-7
schemes were designed; its results are descriptive. The \emph{Held-out Set}
contains 500 decision points from 500 distinct primary-mathematics learners,
opened after the call was frozen and pre-registered. It is disjoint from RQ2,
the earlier design sample, and the Development Set at student and case levels.

Trajectory cases fill a $3\times3$ grid of history length and recent target
accuracy. Their source inputs contain at most 30 preceding attempts, a
knowledge-component summary, the next question's knowledge component as the
goal, and a source-provided prerequisite view. Dialogue cases end at the
student turn, with an inferred goal and no assistant answer. The reference
panel sees neither an action label nor EduGraph state or retrieved context;
this input specification does not establish neutrality across information
conditions. In the Held-out Set, mapping strata are 298 mapped, 116 nearest,
and 86 outside the curriculum. There are 174 weak source-prerequisite signals
and 13 mapped targets with weak curriculum prerequisites (mastery below 0.45);
these are distinct counts whose intersection was not independently re-derived.

The six-action reference vocabulary is \texttt{continue\_explanation},
\texttt{prerequisite\_review}, \texttt{similar\_problem},
\texttt{increase\_difficulty}, \texttt{review}, and
\texttt{diagnostic\_probe}. The corresponding criteria are uncomprehended new
material, a weak prerequisite, insufficient fluency, stable correct work,
a long review gap, and insufficient or contradictory evidence. A wrong answer
alone does not identify which remediation is needed. The runtime action
contract admits nine actions, while its reference policy emits these six;
\texttt{review} has only one reference case in each evaluation set, precluding
a stable per-action interpretation.

\subsection{Models and evaluation references}

The six large backbones are GPT-6.1-Sol, Claude-Opus-5.5, Gemini-4-Argon,
DeepSeek-V4.1-Flash, Qwen3.8-Max, and Gemini-3.8-Flash; the three compact
backbones are GPT-5.4-Nano, Gemini-3.5-Flash-Lite, and Qwen3.8-Flash.
Each runs three seeds at temperature 0.3, giving 27 backbone--seed units.

The five-model prefill panel defined in Section~\ref{sec:s-annotation} uses
plurality with a confidence-based tie-break, rather than a fixed majority
threshold. Evaluated panel members use the leave-one-out plurality of the
other four; non-panel backbones use the full-panel reference. The consolidated
prefills were confirmed by two teachers, but an audit found that some stored
leave-one-out references differ from those final labels. The reported
accuracies retain these model-specific evaluation references and were not
rescored against a common teacher-confirmed reference.

Fleiss $\kappa$ is 0.600 on 459 complete Development cases and 0.505 on 411
complete Held-out cases; mean pairwise accuracy is 0.687 and 0.636. Kimi-K3
failed 265 and 89 requests respectively; these cases use the remaining four
models, without dropping cases or replacing the panel. These statistics
measure prefill agreement, not teacher agreement. A secondary panel shown the
frozen EduGraph block has $\kappa=0.525$ on 380 complete Held-out cases; because
it sees the system's output, it is not independent evidence.

\subsection{Information conditions and results}

Table~\ref{tab:s-e3-inputs} specifies the additional inputs; every condition
receives the current observation and goal. Memory and Memory + EduGraph share
the same raw history, so their comparison measures the added structured
context. The no-history condition is a separate setting.

\begin{table}[t]
\centering
\footnotesize
\setlength{\tabcolsep}{3pt}
\caption{\textbf{RQ3 input conditions.} Additional context beyond the current
observation and goal; the same raw history is used wherever it is included.}
\label{tab:s-e3-inputs}
\begin{tabular*}{\linewidth}{@{}>{\raggedright\arraybackslash}p{0.34\linewidth}>{\raggedright\arraybackslash}p{\dimexpr0.66\linewidth-2\tabcolsep\relax}@{}}
\toprule
Condition & Additional input \\
\midrule
Vanilla & None \\
RAG & Top five EduGraph passages \\
Memory & Raw within-session history \\
EduGraph+State, no raw history & Structured learner state and target prerequisite subgraph \\
Memory + EduGraph & Same raw history, plus the target-centered position card and learner state \\
\bottomrule
\end{tabular*}
\end{table}

\begin{table}[t]
\centering
\footnotesize
\setlength{\tabcolsep}{2.5pt}
\caption{\textbf{Action accuracy under stored evaluation references.}
Development: 724 inspected cases; Held-out: 500 learners evaluated after the
call was frozen. L/S: means over six large/three compact backbones, each with
three seeds. Panel members use leave-one-out references; other models use the
full panel. These references are not uniformly identical to the final
teacher-confirmed labels. \na{} = condition not run on that split.}
\label{tab:s-e3-core}
\begin{tabular*}{\linewidth}{>{\raggedright\arraybackslash}p{\dimexpr0.48\linewidth-2\tabcolsep\relax}>{\centering\arraybackslash}p{\dimexpr0.13\linewidth-2\tabcolsep\relax}>{\centering\arraybackslash}p{\dimexpr0.13\linewidth-2\tabcolsep\relax}>{\centering\arraybackslash}p{\dimexpr0.13\linewidth-2\tabcolsep\relax}>{\centering\arraybackslash}p{\dimexpr0.13\linewidth-2\tabcolsep\relax}}
\toprule
 & \multicolumn{2}{c}{Development} & \multicolumn{2}{c}{Held-out} \\
\cmidrule(lr){2-3}\cmidrule(lr){4-5}
Information given to the agent & L & S & L & S \\
\midrule
Observation only (Vanilla) & 0.512 & 0.471 & \na & \na \\
\rowcolor{bestblue} + EduGraph passages (RAG) & 0.516 & 0.481 & \na & \na \\
History (Memory) & 0.698 & 0.577 & 0.632 & 0.482 \\
\rowcolor{bestblue} EduGraph+State, no raw history & 0.671 & 0.613 & \na & \na \\
\rowcolor{bestblue} Memory + EduGraph & \textbf{0.721} & \textbf{0.646} & \textbf{0.659} & \textbf{0.570} \\
\bottomrule
\end{tabular*}
\end{table}

Table~\ref{tab:s-e3-core} complements the main paper's per-backbone comparison
with all core conditions. Raw history improves Development accuracy over
Vanilla by 18.6 points for large models and 10.5 for compact models; RAG changes
it little. Without raw history, EduGraph+State is 2.7 points below Memory for
large models (McNemar $p=0.143$) and 3.7 above it for compact models; this
condition was not run on the Held-out Set.

Adding EduGraph to the same history improves large/compact accuracy by
2.3/6.9 points on Development (95\% CIs 0.6--3.9/5.1--8.6) and 2.7/8.8 on
Held-out (0.6--5.0/6.3--11.2). The Held-out compact-model gain has exact
McNemar $p<0.001$; the large-model gain is bootstrap-positive but not
exact-McNemar significant ($p=0.076$). Eight of nine backbones improve on each
set, while Gemini-4-Argon declines. Thus the cohort means summarize
heterogeneous effects under the stored references. Agreement on fixed inputs
does not establish how learners would respond to the selected actions or
demonstrate learning gains.

\end{document}